\documentclass[lettersize,journal]{IEEEtran}
\usepackage{amsmath,amsfonts}
\usepackage{algorithmic}
\usepackage{algorithm}
\usepackage{array}
\usepackage{textcomp}
\usepackage{stfloats}
\usepackage{verbatim}
\usepackage{graphicx}
\usepackage{cite}

\usepackage{times}
\usepackage{color}
\usepackage{multirow}
\usepackage{framed}
\usepackage{comment}
\usepackage{booktabs} % To thicken table lines
\usepackage{algorithm}
\usepackage{algorithmic}
\usepackage{url,hyperref}
\usepackage[OT1]{fontenc}

\newcommand{\mycite}[1]{\cite{#1}}

\def\hcolor{black}
\def\hcolorb{black}

\makeatletter
\let\MYcaption\@makecaption
\makeatother

\usepackage[font=footnotesize]{subcaption}

\makeatletter
\let\@makecaption\MYcaption
\makeatother
\begin{document}
\bstctlcite{IEEEexample:BSTcontrol}

\title{Foundational Policy Acquisition via \\ Multitask Learning for Motor Skill Generation}

\author{Satoshi Yamamori and Jun Morimoto
\thanks{Satoshi Yamamori and Jun Morimoto are with Learning Machines Group, the Graduate School of Informatics, Kyoto University, Kyoto, 606-8501, Japan, and with the Department of Brain Robot Interface (BRI), Brain Information Communication Research Laboratory Group (BICR), Advanced Telecommunications Research Institute International (ATR), Kyoto, 619-0288, Japan.
(E-mail: morimoto@i.kyoto-u.ac.jp, yamamori@atr.jp)

This is an Accepted Manuscript of an article published by IEEE Transactions on Cognitive and Developmental Systems on 19 February 2025, available at: \url{https://doi.org/10.1109/TCDS.2025.3543350}.
% \copyright 2025 IEEE.  Personal use of this material is permitted.  Permission from IEEE must be obtained for all other uses, in any current or future media, including reprinting/republishing this material for advertising or promotional purposes, creating new collective works, for resale or redistribution to servers or lists, or reuse of any copyrighted component of this work in other works.
}
}

% The paper headers
\markboth{IEEE TRANSACTIONS ON COGNITIVE AND DEVELOPMENTAL SYSTEMS, VOL. X, ISSUE X, 2025}%
{Shell \MakeLowercase{\textit{et al.}}: A Sample Article Using IEEEtran.cls for IEEE Journals}

% \IEEEpubid{0000--0000/00\$00.00~\copyright~2025 IEEE}

\maketitle

%Abstract
\begin{abstract}
In this study, we propose a multitask reinforcement learning algorithm for foundational policy acquisition to generate novel motor skills. \textcolor{\hcolor}{Learning the rich representation of the multitask policy is a challenge in dynamic movement generation tasks because the policy needs to cope with changes in goals or environments with different reward functions or physical parameters. Inspired by human sensorimotor adaptation mechanisms, we developed the learning pipeline to construct the encoder-decoder networks and network selection to facilitate foundational policy acquisition under multiple situations. First, we compared the proposed method with previous multitask reinforcement learning methods in the standard multi-locomotion tasks. The results showed that the proposed approach outperformed the baseline methods. Then, we applied the proposed method to the ball heading task using a monopod robot model to evaluate skill generation performance. The results showed that the proposed method was able to adapt to novel target positions or inexperienced ball restitution coefficients but to acquire a foundational policy network, originally learned for heading motion, which can generate an entirely new overhead kicking skill.}

\end{abstract}
%%%%%%%%%%%

\begin{IEEEkeywords}
  Robot learning; Multitask reinforcement learning; Domain adaptation; Policy selection
\end{IEEEkeywords}

\section{Introduction}

\IEEEPARstart{M}{ultitask} reinforcement learning (multitask RL) has been gaining attention as a promising tool to develop human-level skillful agents.
\textcolor{\hcolor}{In particular, the meta-reinforcement learning (meta-RL) approach that updates shared policy parameters among different policies for different tasks has become an important research topic for optimizing the parameters of the policies to cope with the multitask RL problems \mycite{clavera2019, rakelly2019}.
Brain-inspired frameworks for multitask learning problems have also been explored \cite{chen2022,jin2024}.
In this study, we propose adopting the idea of multitask RL to acquire a foundational motor skill that can generate novel motor skills by using encoder-decoder networks.} As the computational neuroscience studies stated \cite{Krakauer2019}, \cite{Morehead2021}, \cite{Tsay2022}, human sensorimotor adaptation mechanisms may contain explicit motor skill learning processes and non-intentional implicit adaptation processes to circumstances. 
Inspired by these sensorimotor adaptation systems, we propose using a multitask RL framework to acquire a foundational policy that can take the role of implicit adaptation and optimize latent variables of the decoder network to generate novel skills as explicit motor learning.

\textcolor{\hcolor}{
First, to evaluate the multitask learning performance, we compared the proposed method with previous meta-RL methods in the standard multi-locomotion benchmarks \cite{finn2017,rakelly2019}. 
Then, we applied the proposed method to the ball heading task using a monopod robot model to evaluate skill generation performance} since
football is one of the most popular sports in the world \cite{Krakauer2019}.
Furthermore, for example, the ball can fly in completely different directions with a similar heading motion by slightly changing the hitting position, the force applied to the ball, or when the friction of the ball is different.
In such cases, continuous task changes may have infinite numbers of task goals, and preparing policy networks for all these tasks becomes impossible.

\begin{figure}[tbp]
    \centering
    \includegraphics[width=0.99\linewidth]{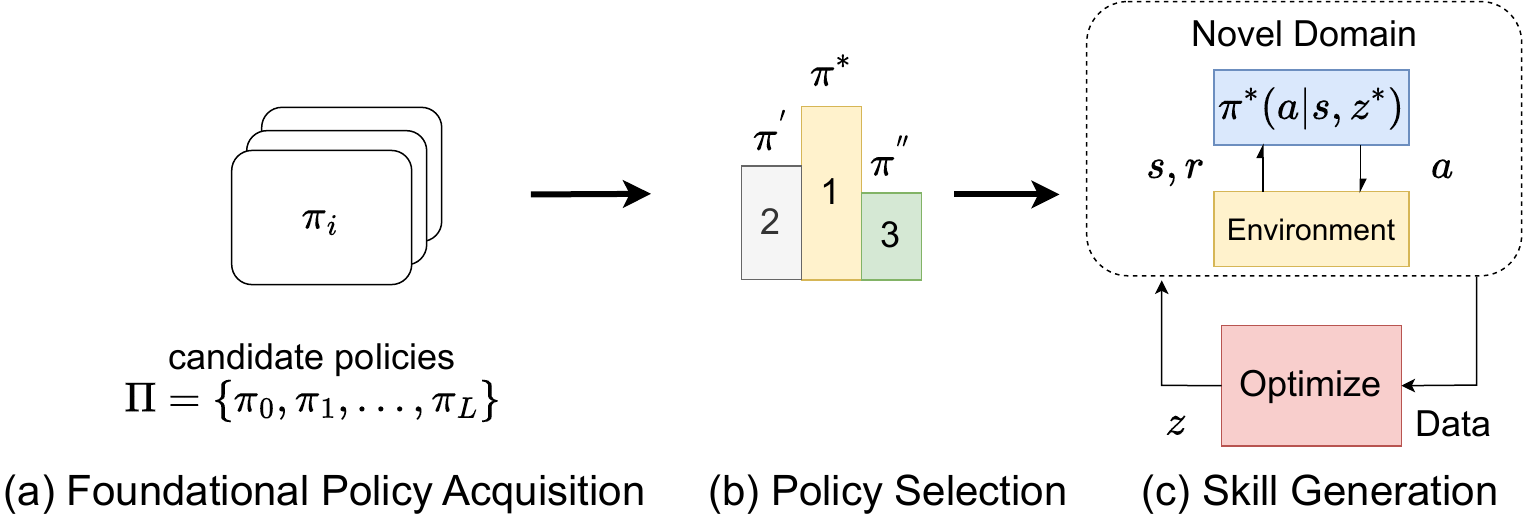}
    \caption[snapshot]{
        Proposed three-phase multitask learning method.
        Proposed pipeline comprises three phases: acquisition, selection, and generation.
        In the foundational policy acquisition phase, multiple candidate policies are generated under multiple tasks.
        (a) $i$-th candidate policy $\pi_i$ is trained under multiple tasks.
        (b) Policy selection phase chooses a policy from the policy set $\Pi$ based on the performance index defined in Eq. (\ref{eq:policy-selection-index}).
        (c) Context-related latent variable $z$ was estimated to generate skills to handle the novel tasks: unknown rewards and environmental settings.
        We showed that the Bayesian optimization method can be efficiently used to update the latent variable in the skill generation.
    }
    \label{fig:PolicySelectionConcept}
\end{figure}

\textcolor{\hcolor}{As illustrated in Fig. \ref{fig:PolicySelectionConcept}, this study proposes a three-phase training procedure for policies represented by deep neural networks to cope with the multitask RL problems with the task changes. 
Unlike the previous meta-RL methods, our proposed method can generate a novel movement skill based on a newly developed variational inference framework to train the policy network models.}
\begin{figure}[tbp]
    \centering
    \begin{minipage}{0.95\linewidth}
        \subfloat[][Target direction]{
            \includegraphics[width=0.45\textwidth]{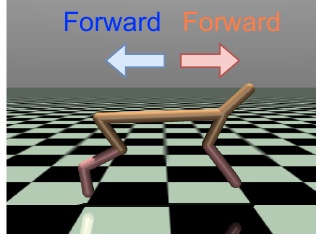}
            \label{fig:half-cheetah-dir-snap}
        }    
        \subfloat[][Target veloctiy]{
            \includegraphics[width=0.45\textwidth]{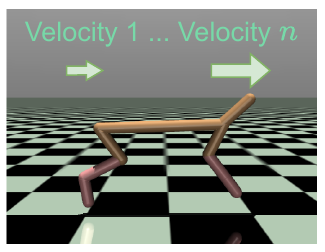}
            \label{fig:half-cheetah-vel-snap}
        }
    \end{minipage}
    \begin{minipage}{0.99\linewidth}
        \subfloat[][Multitask heading]{
            \includegraphics[width=0.3\textwidth]{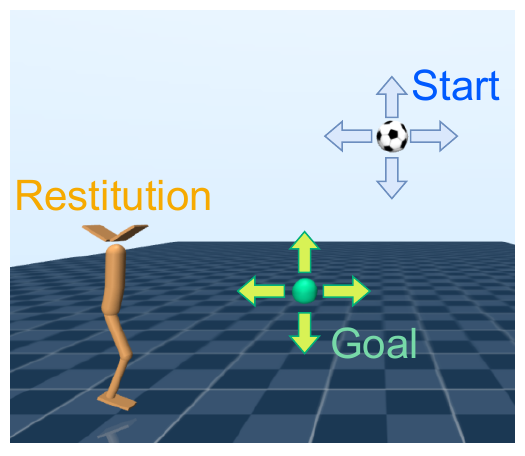}
            \label{fig:SanpshotTask}
        }    
        \subfloat[][Throw in]{
            \includegraphics[width=0.3\textwidth]{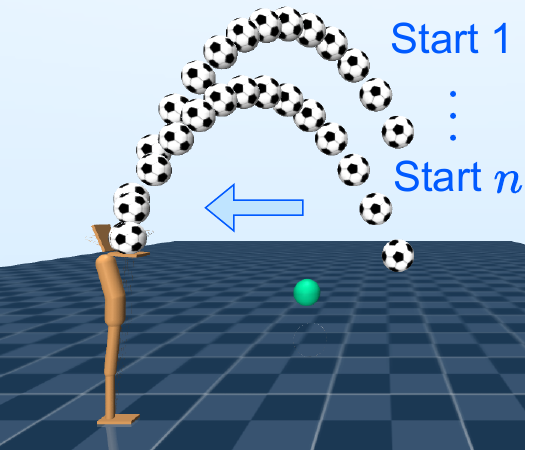}
            \label{fig:SnapshotBeforeHit}
        }
        \subfloat[][Hitting ball]{
            \includegraphics[width=0.3\textwidth]{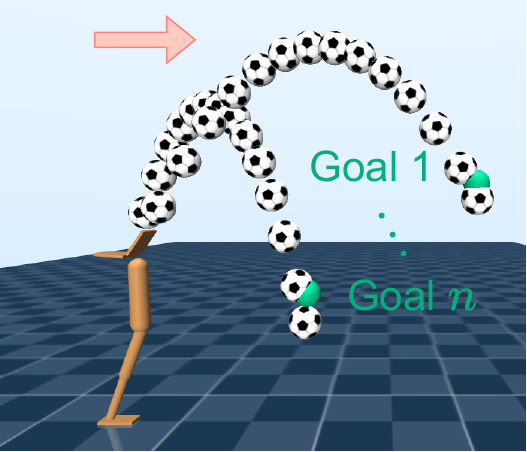}
            \label{fig:SnapshotAfterHit}
        }
    \end{minipage}
    \caption[snapshot]{
        \textcolor{\hcolor}{Multitask environments to generate a variety of dynamic movements.
        (a) Half-Cheetah-Dir domain. Goal is to move forward or backward as fast as possible.
        (b) Half-Cheetah-Vel domain. Goal is to reach a target velocity  \cite{finn2017,rakelly2019}}.
        (c, d, e) Implicit multitask heading. Start position, goal position, and the coefficient of the restitution are varied.
    }
    \label{fig:snapshots-adapted}
\end{figure}
\IEEEpubidadjcol
\textcolor{\hcolor}{We first evaluated our proposed approach in the standard multi-locomotion benchmarks illustrated in Figs. \ref{fig:snapshots-adapted}(a), (b). In the Half-Cheetah-Dir domain, the goal is to move forward or backward as fast as possible with using the Half-Cheetah model. In the Half-Cheeetah-Vel domain, the goal is to reach a target velocity \cite{finn2017,rakelly2019}. 
We then evaluated the skill generation performance of our proposed policy adaptation method using the monopod motor control task.}
In this task, as depicted in Figs. \ref{fig:snapshots-adapted}(c), (d), (e), the monopod model tries to hit the ball back to the target point while maintaining the posture until it hits the ball.  
To show the skill generation performance of the proposed method, the target points, as well as the coefficient of restitution that defines the physical interaction between the ball and the monopod model, were changed.
We demonstrated that the monopod model was successfully adapted to the implicitly changing task settings. Furthermore, we showed that the new movement skill could be generated by using the acquired foundational model.
The contributions of our study can be summarized as follows:
\begin{itemize}
    \item We developed a policy adaptation pipeline that can cope with implicit multitask problems. The proposed method is composed of three phases: 1) foundational policy acquisition, 2) policy selection, and 3) skill generation, as shown in Fig. \ref{fig:PolicySelectionConcept}.
    \item We demonstrated that selecting the base policy from the policy pool derived in the foundational policy acquisition phase in terms of learning performance is the key to successful skill generation. \textcolor{\hcolor}{To evaluate the multitask learning performance, we compared our proposed method with previous meta-RL methods in the standard meta-locomotion domains.
    The results showed that the proposed approach outperformed the baseline methods. }
    \item We showed that the monopod model was able to hit the ball back to novel target points with different physical interaction settings using the proposed skill generation method. \textcolor{\hcolor}{We further demonstrated that the monopod model was able to generate a novel kicking skill that was not explicitly acquired in the foundational policy acquisition phase.}
\end{itemize}

The remainder of this paper is organized as follows:
In section I\hspace{-.1em}I, we described the related studies.
In section I\hspace{-.1em}I\hspace{-.1em}I, we introduced the proposed adaptation method.
In section I\hspace{-.1em}\nolinebreak V, we explained the experimental settings for the \textcolor{\hcolor}{Half-Cheetah and} monopod model used for evaluating the proposed method.
In section V, we presented the results.
In section V\hspace{-.1em}I,  we discussed the comparison of the proposed method with the standard approach as well as the limitations of the proposed method.
Finally, in Section V\hspace{-.1em}I\hspace{-.1em}I, we presented the conclusions of the study.
\section{Related Works}
\subsection{Policy Learning Methods for Multiple Task Goals}
To acquire a policy that can cope with multiple task goals, goal-conditioned RL methods have been developed \mycite{andrychowicz2017},\mycite{schaul2015},\mycite{liu2022},\mycite{whitehead1993}, \mycite{singh1991}.
\textcolor{\hcolor}{
Goal-conditioned approaches have also been explored in imitation learning frameworks \mycite{ude2010,wu2022}.
By acquiring a goal-conditioned value function, the learned policy can generate optimized behaviors according to the continuously changing task goals \mycite{schaul2015,nair30}, \mycite{foster2002}.
Moreover, some studies proposed to use an image as a task goal \mycite{liu2022,nair30}.
However, these approaches essentially select a policy corresponding to a specified task goal and cannot handle implicitly changing task goals.
This study proposes a three-phase learning and adaptation procedure for coping with implicitly changing environmental settings and task goals. 
As a part of the proposed pipeline, we considered policy selection problem \mycite{gulcehre2020} to derive a foundation policy. 
Concretely, the policy selection phase chooses a policy
from a policy set based on the performance index defined by the Kullback--Leibler (KL) divergence between trajectory distributions (see Section \ref{sec:cMDP} for details).}

\subsection{Data-driven System Identification}
The online data-driven approach is used to cope with a novel environment, and adaptive control methods have been widely studied \mycite{slotine1991applied}.
If we cannot rely on a parametric environmental model, the model is usually built based on domain-specific knowledge.
In that case, we can use an approximated model, possibly represented by a neural network, to bridge the gap between the current estimated and the actual environmental settings.
Our early studies proposed collecting data by first using a robust policy that can deal with the novel environment and further improving the policy based on the approximated physical model by using the newly acquired data from the novel environment \mycite{morimoto2002, morimoto2009}.
Recently, instead of using the robust policy, acquiring a policy network that takes physical parameters as inputs has been proposed \mycite{yu2017}. The environment-aware policy was acquired under various environmental settings in the training phase. In the test phase, the physical parameters were identified to improve the policy performance. This approach has been further explored using a low-dimensional representation of the physical parameters. In the training phase, the low-dimensional latent parameter space was extracted. Then, in the test phase, the latent parameters were identified through interaction with the novel environment.
The above approach was adopted to deal with the simulation-to-real gap \mycite{yu2019a, peng2020}. In this study, we applied these system identification approaches to the implicit multitask RL problem. 
We showed that the policies can be adapted not only to novel environments but also to novel tasks. 

\subsection{Derivative-free Optimization}
For solving optimization problems that have many low-quality local minima, using a gradient-based optimization method is unsuitable.
A derivative-free approach such as the evolution strategy can be helpful in such cases.
Furthermore, the evolution strategy has been explored for solving optimization problems \mycite{wierstra2014,salimans2017}.
In the adaptation phase of our proposed method, we adopted the derivative-free RL approach to avoid low-quality local minima in the 
context-related variable space extracted by the acquired encoder network.
As the derivative-free optimization method, we used the tree-structured Parzen estimator (TPE) \mycite{bergstra2011, bergstra2013}, which is well known as the high-performance parameter optimization method.
In general, derivative-free methods do not work well in high-dimensional optimization problems.
However, since the context-related variable space extracted in the adaptation phase was low-dimensional, our proposed method performed well with the derivative-free optimization method. 

\section{Method}
In this section, we first defined contextual MDP (cMDP) as multitask reinforcement learning, in which the task goal or environments change implicitly.
We then introduced the novel variational inference-based reinforcement learning method to cope with cMDP.

\subsection{Solving MDP through KL-divergence Minimization}
Here, we considered a cumulative reward to search for the optimal policy in the maximum entropy RL framework as follows:
\begin{equation}
    \max_{\pi} \ \mathbb{E}_{\tau|\pi} \left [ \sum_{t=0}^T  r(s_t, a_t) -\alpha \log \pi(a_t|s_t) \right ], 
    \label{eq:ConditionalExpectationCumreward}
\end{equation}
where $\tau=\{s_0, a_0, ..., s_t, a_t,...,s_T, a_T\}$ is the state and action trajectory; $r(s_t,a_t)$ is the reward function; $\pi(a_t|s_t)$ is the policy; and the hyperparameter $\alpha \geq 0$ regulates the trade-off between exploration and exploitation\mycite{haarnoja2018a}.
It is known that the entropy-regularized RL problem can be reformulated as the inference problem of the policy distribution, where policy distribution is derived by minimizing the KL divergence between the distributions of the trajectory generated by the policy as,
\begin{equation}
    q(\tau) = \prod_{t=0}^{T}q(s_{t+1}|s_t, a_t) \pi(a_t|s_t), 
\end{equation}
and the trajectory distribution induced by the reward as,
\begin{equation}
    p(\tau) = \prod_{t=0}^{T}p(s_{t+1}|s_t, a_t)\exp(r(s_t, a_t) / \alpha).
\end{equation}
We then found that minimizing the KL divergence leads to the maximization of the entropy-regularized cumulative reward as follows:
\begin{align}
     \alpha \mathrm{KL}[q|p] &= \alpha \mathbb{E}_q [\log q - \log p]  \nonumber \\
    &= \mathbb{E}_q \left[\sum_{t=0}^T \alpha \log \pi_t - r_t \right].
\end{align}
Notably, we can regard the above approach as discounted cumulative reward optimization by simply considering modified dynamics ${\bar p}(s_{t+1}|s_t,a_t)=\gamma p(s_{t+1}|s_t,a_t)$. 
In this case, transition with $1-\gamma$ into an absorbing state with zero rewards no matter what actions were taken \mycite{levine2018}.

\subsection{Solving cMDP}
\label{sec:cMDP}
This subsection introduced the proposed variational inference-based RL method for solving cMDP.
Contextual MDP $\{\mathcal{A},\mathcal{S}, \mathcal{C}, r,p, p_0\}$ has the state space $\mathcal{S}$, action space $\mathcal{A}$, context space $\mathcal{C}$, reward function $r(s, a, c)\in \mathbb{R}$, state transition probability $p(s'|s,a,c)\in \mathbb{R}$, and initial distribution of the state and context $p_0 = p(s_0)p(c) \in \mathbb{R}$.
State transition probability and reward functions are conditioned with a vector $c$ in a context space $\mathcal{C}$.
To acquire the optimal policy in cMDPs, we trained a policy $\pi(a|s,z)$, which depended on states $s$ and the latent variable $z$.
Here, the latent variable $z$ is a context-relevant feature vector sampled from an encoder distribution $q(z|c)$.
In other words, the encoder converts the current context variables $c$ to the latent variable $z$.

In this study, we proposed a learning procedure to optimize the policy and encoder for a cMDP.
\begin{align}
    \min_{\pi, q}\ & \alpha \mathrm{KL}[q|p] = \alpha \mathbb{E}_q \left [\log q - \log p \right ], \label{eq:VI_KL}\\
    q(\tau, z, c) &= p(c)q(z|c)\prod_{t=0}^{T}p(s_{t+1}|s_t, a_t, c) \pi(a_t|s_t, z), \nonumber \\
    p(\tau, z, c) & = p(c)\rho(z)\prod_{t=0}^{T}p(s_{t+1}|s_t, a_t, c)\exp(r(s_t, a_t, c) / \alpha) \nonumber.
\end{align}
The variational distribution $q$ is the probability of trajectory $\tau$ given context $c\sim p(c)$ and latent variable $z \sim q(z|c)$.
Likewise, a generative model $p$ is the target distribution close to the variational distribution $q$, and $\rho(z)$ is the prior distribution.
The variational distribution and generative model share the context distribution $p(c)$ and transition distribution $p(s_{t+1}|s_t, a_t, c)$.
Therefore, cumulative rewards with entropy regularization by canceling out shared distribution were obtained as follows:
\begin{align}
     \alpha \mathrm{KL}[q|p] &= \alpha \mathbb{E}_q [\log q - \log p]  \label{eq:original-kl-divergence} \\
    &= \mathbb{E}_q \left[\alpha\log \frac{q(z|c)}{\rho(z)} + \sum_{t=0}^T \alpha \log \pi_t - r_t \right] + \mathrm{const}\nonumber \\
    &\propto \mathbb{E}_{s_0,c} \left[ \alpha\mathrm{KL}\left [ q(z|c)|\rho(z) \right] - V^\pi(s_0, c) \right ] , \label{eq:ValueKL}\\
    & V^\pi(s, c) := \mathbb{E}_{\tau, z|s, c} \left[ \sum_{t=0}^T r_t - \alpha\log \pi_t  \right]. \label{eq:ValueDefinition}
\end{align}
In the policy evaluation step, the value function was computed by solving the soft Bellman equation as follows:
\begin{align}
    Q^\pi(s, a,c) &:= 
        r(s, a, c) + \mathbb{E}_{p(s'|s, a, c)} 
        [  V^\pi(s', c) ] 
        \label{eq:DomainSoftBellmanQEquation}, \\ 
    V^\pi(s, c) &= 
        \mathbb{E}_{q(z|c)\pi(a|s, z)} 
        [ Q^\pi - \alpha \log \pi ].
        \label{eq:DomainSoftBellmanEquation}
\end{align}
In the policy improvement step, the policy and encoder were updated to minimize the KL divergence as,
\begin{align}
	&\min_{q,\pi} \mathbb{E}_{s, c}\left [ \beta \mathrm{KL}\left[q(z|c) \vert \rho(z) \right] - V^\pi(s,c) \right]
     . \label{eq:EncoderKLDivergence}
\end{align}
By comparing Eq. (\ref{eq:original-kl-divergence}) with Eq. (\ref{eq:EncoderKLDivergence}),
the policy and encoder minimize the KL divergence between the variational distribution and the generative model.

\subsection{Learning Algorithm}
\begin{figure*}[tbp]
    \centering
    \begin{minipage}{0.8\linewidth}
        %\vspace{-12pt}
        % \addtocounter{subfigure}{3}
        \subfloat[h][Foundational policy acquisition]{
            \includegraphics[width=0.5\textwidth]{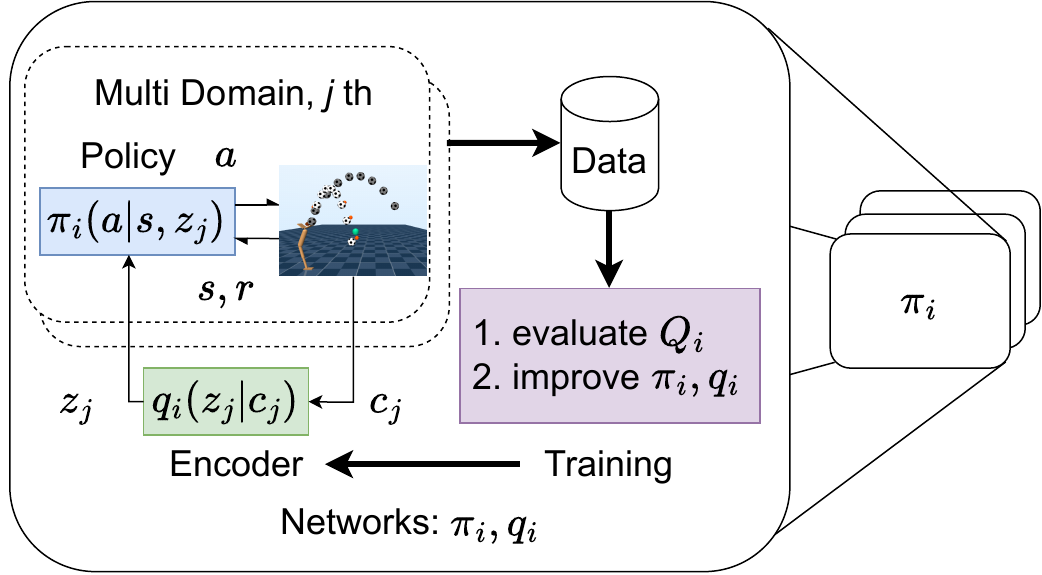}
            \label{fig:policy-acquisition}
        }
        \subfloat[h][Policy selection]{
            \includegraphics[width=0.5\textwidth]{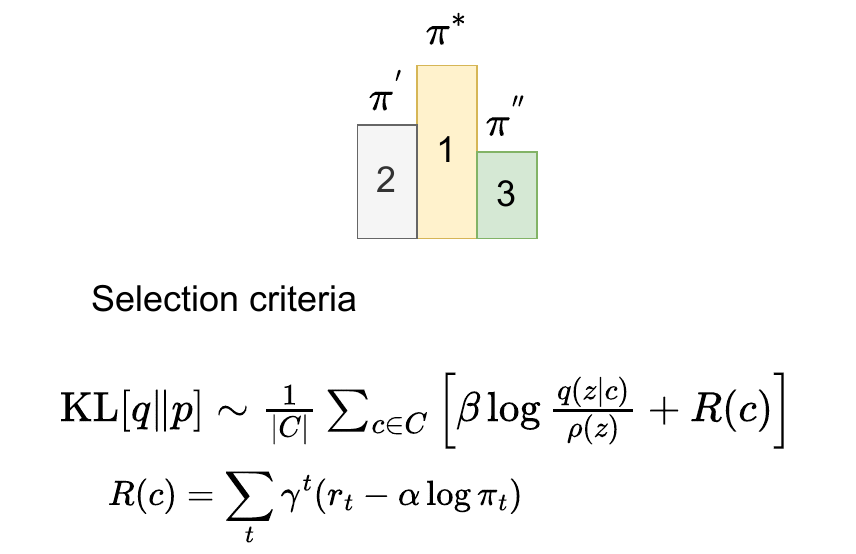}
            \label{fig:policy-selection}
        }
    \end{minipage}
    \caption[snapshot]{
        \textcolor{\hcolor}{
        Foundational policy acquisition and policy selection.
        We demonstrated that selecting a foundational policy model according to the learning performance with randomized task parameters is the key to successful adaptation to multiple reward and environmental settings. Furthermore, our proposed method simultaneously acquires the optimal policy for each context and encoder to embed context variables that correspond to the task settings.}
    }
    \label{fig:Concept-detail}
\end{figure*}

The proposed multitask learning algorithm is composed of three phases: acquisition, selection, and generation phases,
as shown in Figs. \ref{fig:PolicySelectionConcept} \textcolor{\hcolor}{and \ref{fig:Concept-detail}}.

\subsubsection{Foundation Policy Acquisition}
\label{sec:LearningAlgorithm}
\begin{figure}[t]
    \includegraphics[width=0.99\linewidth]{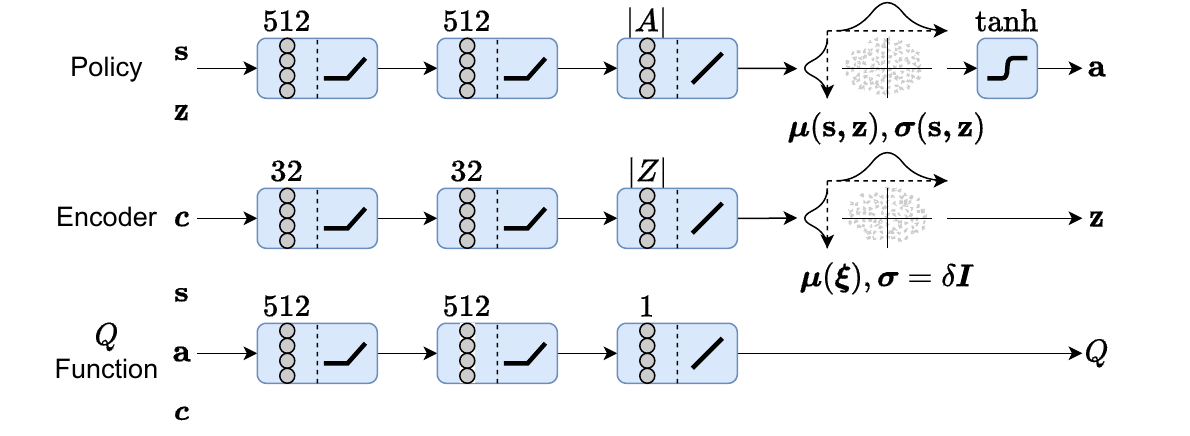}
    \centering
    \caption[Network Architecture]{Network architecture. Our proposed method utilized three neural networks, namely, policy network, encoder network, and $Q$-network which approximate the action-value function. 
    All networks contain two hidden layers, 
    and the numbers on each block represent the layer size.
    Inputs to the policy network included state $s$ and latent variable $z$.
    The encoder network considered the context variable $c$ as the input.
    The $Q$-network recieved state $s$, action $a$, and context $c$ as inputs.}
    \label{fig:NetworkArch}
\end{figure}

In the foundation-policy acquisition phase, the policy was trained under multiple contexts (see Algorithm \ref{alg:DomainRandomization}).
In the policy evaluation step, we first updated the approximated action value function $Q_{\phi}$ to reduce the Bellman residual as follows:
\begin{equation}
    \min_{\phi} \mathbb{E}_{\mathcal{D}}
        \left ( Q(s, a, c; \phi) - r(s, a, c) - \gamma \bar{V}(s', c) \right )^2.
\end{equation}
We then updated the policy and the encoder using the KL divergence introduced in Eq. (\ref{eq:EncoderKLDivergence}) as follows:
\begin{equation}
    \max_{\psi, \eta}  \mathbb{E}_{(s, c)\sim \mathcal{D}} [\bar{V}(s, c)]  - \beta \mathbb{E}_{c\sim \mathcal{D}} \mathrm{KL}\left[ q \| \rho \right ],
        \label{eq:FittedQ}
\end{equation}
where $\mathcal{D}$ denotes the sample set from the replay buffer, and the state value function $V$ can be derived as follows:
\begin{align}
    &\bar{V}(s, c) = Q(s, a', c;\phi) - \alpha \log \pi(a'|s, z';\psi),\\
    & a' \sim \pi(a|s, z';\psi), \quad z'\sim q(z|c;\eta). \nonumber
\end{align}
Here, $\phi$, $\psi$, and $\eta$ represent the parameters of the Q function, the policy, and the encoder, respectively.
Fig. \ref{fig:NetworkArch} shows the architecture of each network.
The prior of the latent variable $\rho(z)$ was derived by a standard multivariate normal distribution.
The replay buffer $\mathcal{D}$ stored the tuple $\{s, a, r, s', c \}$.

\begin{algorithm}
\caption{Foundation Policy Acquisition}
\label{alg:DomainRandomization}
\begin{algorithmic}[1]
    \STATE \textbf{Input}: $\phi_{\{1, 2\}}, \psi, \eta$
    \STATE $\phi_{\{1, 2\}}^\mathrm{target} \leftarrow \phi_{\{1, 2\}}$
    \STATE $\mathcal{D} \leftarrow \emptyset$
    \FOR {each epoch}
        \STATE sample domain and encoded parameter: \\
        $\qquad z\sim q(z|c;\eta), c\sim p(c)$
        \FOR{collect state-action steps}
            \STATE $a_t\sim \pi(a|s,z;\psi)$
            \STATE $s_{t+1}\sim p(s_{t+1}|s_t,a_t, c)$
            \STATE $\mathcal{D}\leftarrow \mathcal{D}\cup\{ (s_t, a_t,r(s_t, a_t,c), s_{t+1}, c)\} $
        \ENDFOR
        \FOR{update iterations}
            \STATE update policy $\psi$ 
            \STATE update encoder $\eta$
            \STATE update $Q$-function $\phi_{1, 2}$ 
            \STATE update temperature $\alpha$ 
            \STATE update target $Q$-function $\phi_{1, 2}^\mathrm{target}$ 
        \ENDFOR
    \ENDFOR
    \STATE \textbf{Output } policy and encoder network weights $\psi, \eta$
\end{algorithmic}
\end{algorithm}

\subsubsection{Selection}
One key idea of our proposed method was to select a policy model based on the return values in the foundational policy acquisition phase. As a well-known issue, the task performance of a policy highly depends on an exploration procedure determined by a random seed. Thus, in this paper, we proposed the acquisition of multiple policy models in the offline RL manner to find a high-quality policy that can adapt to a novel task from the foundational policy models.
\color{\hcolor}
The detailed selection procedure is explained in Algorithm \ref{alg:PolicySelection}.
The $k$-th candidate policy collects the sampled return value $R(c)$ for each context from a context set.
The performance index $l_k$ defined in Eq. (\ref{eq:EncoderKLDivergence}) is stored with the candidate policy $\pi_k$ for policy selection as follows:
\begin{align}
    l_k &= \frac{1}{|\mathcal{C}|} \sum_{c\in \mathcal{C}} \left [ \beta \log \frac{q(z|c)}{\rho(z)} +  R(c) \right ],
    \label{eq:policy-selection-index}\\
    R(c) &=  \sum_t \gamma ^t \left( r_t - \alpha \log \pi_t(z) \right), z\sim q(z|c). \nonumber
\end{align}
We then select the policy with the highest performance index.
\color{black}
\begin{algorithm}
\caption{Selection Phase}
\label{alg:PolicySelection}
\begin{algorithmic}[1]
    \STATE Reset candidate policies $\Pi = \phi$
    \STATE \textcolor{\hcolor}{Sample context variables to define context set $\mathcal{C}$}
    \FOR{iteration $k=1,\dots,K$}
        \STATE Train $\{\pi_k, q(z|c)\}$ $\triangleright$ Foundational Policy Acquisition
        \STATE Rollout $\{\pi_k, q(z|c)\}$ and collect \\
            $R(c) := \sum_t {\gamma^t (r_t - \alpha \log \pi_t)}$ for each $c \in \mathcal{C}$
        \STATE \textcolor{\hcolor}{Evaluated policy selection index $l_k$ in Eq. (\ref{eq:policy-selection-index})}
        \STATE $\Pi \leftarrow \Pi \cup \{\pi_k, l_k\}$
    \ENDFOR
    \STATE Select $\pi_s$ based on sorted $\Pi$ by $\{l_k\}$
    \STATE Deploy $\pi_s$ to domain with new context $c'$
\end{algorithmic}
\end{algorithm}

\subsubsection{Skill Generation}
\label{sec:Adaptation_Phase}
The encoder network learns the latent variable space, where the context variables are embedded in this latent space.
In the skill-generation phase, we only optimized the latent variable so that the policy could cope with a novel domain. 

To estimate the latent variable $z$ based on the KL loss under the context $c$, we solve the following optimization problem:
\begin{align}
    z^{*}&= \underset{z}{\arg \max}\ J(z) \nonumber,\\
    J(z) &=  R(z) - \beta \mathrm{KL}[P(*|z)\|\rho(*)], \label{eq:objectEncodeParameter}\\
    &R(z) = \sum_{t=0}^{T-1} \gamma^{t} (r_t-\alpha \log \pi_t(z)). \nonumber
\end{align}
Skill-generation encoder $P(*|z)$ is represented with multivariate normalized distribution $\mathcal{N}(z, \mathbf{I})$.
We adopted black-box optimization methods to explore a one-dimensional latent space.
As a concrete optimization method, we chose the TPE, which has been used as a standard tool for searching hyperparameters \mycite{bergstra2011}, because it shows empirically better optimization performance than other types of Bayesian optimization methods, such as the GPs and random search methods \mycite{bergstra2011, bergstra2013}.
To solve the optimization problem, TPE estimates the conditional distribution $p(z|J)$ from sampled data $\mathcal{D}=\{(z, J)\}$.
We adopted the off-the-shelf implementation in the hyperparameter optimization framework, Optuna \mycite{akiba2019}.
The mean $z$ of encoder distribution is suggested by TPE for each iteration, as shown in Algorithm \ref{alg:DomainAdaptation}.
\textcolor{\hcolor}{The algorithm collects rewards series from the deterministic action $a=\mu(s, z_k)$, which is the pre-layer of noise injection in Fig. \ref{fig:NetworkArch} based on the suggested latent variable $z_k$ to estimate the return value $J(z_k)$ and store it in the replay buffer.}
The TPE updates the posterior distribution model for the next suggestion.
The rollout will be iterated $k_\mathrm{max}$ times.
Finally, the algorithm returns the best value $z^*$ stored in the replay buffer $\mathcal{D}$.
\begin{algorithm}
\caption{Skill Generation Phase}
\label{alg:DomainAdaptation}
\begin{algorithmic}[1]
    \STATE $\pi \leftarrow $ selected policy in Selection Phase
    \STATE $\mathcal{D} \leftarrow \emptyset$
    \STATE TPE model initialization
    \FOR{iteration $k =1, \dots, k_\mathrm{max} $}
        \STATE $z_k \leftarrow$ TPE suggests
        \STATE \textcolor{\hcolor}{$J(z_k) \leftarrow$ Rollout policy $a=\mu(s, z_k)$ in Eq. (\ref{eq:objectEncodeParameter})}
        \STATE Store $(z_k,J(z_k))$ in $D$
        \STATE TPE updates model $p(z|J)$
    \ENDFOR
    \STATE Select $z^*$ based on the sorted $\mathcal{D}$ by indices $J_k$
\end{algorithmic}
\end{algorithm}

\subsubsection{Network models}
\textcolor{\hcolor}{
The policy network was implemented using the multilayer neural network model.
The network model had two hidden layers, shown in Fig. \ref{fig:NetworkArch}.
The input variables to the network included the robot state $s$ and the latent variable of the encoder $z$.
The encoder network embedded the context variable $c$ to the latent variable $z$ and implemented a neural network model with two hidden layers.
In the foundational policy acquisition phase, the encoder and Q-function networks were trained using actual context variables $c$. 
Subsequently, in the skill generation phase, the encoder network was not used. The latent variable $z$ was estimated through adaptation trials.
}
\section{Experiments}
\color{\hcolor}
To evaluate the multitask learning performance, we first compared the proposed framework with previous meta-RL methods in the standard multi-locomotion domains \cite{finn2017,rakelly2019} using the Half-Cheetah model. We then applied the proposed method to the ball heading task using a monopod robot model to evaluate skill generation performance. We adopted a dynamics simulator MuJoCo \mycite{todorov2012} for these evaluations.

\subsection{Multi-locomotion Domains}
We compared our proposed approach with previous methods in the Half-Cheetah-Dir and Half-Cheetah-Vel domains. In the Half-Cheetah-Dir domain, the goal is to move forward or backward as fast as possible. In the Half-Cheetah-Vel domain, the goal is to reach a target velocity \cite{finn2017,rakelly2019}.
In the policy acquisition phase, one of the target directions was selected as a context value in the Half-Cheetah-Dir domain, and the target velocity was sampled from a uniform distribution with a range of $0.0$ to $3.0$ m/s \cite{rakelly2019}  as a context value in the Half-Cheetah-Vel domain. These context values are fixed for the duration of one learning episode, where the learning agents collected data for $200$ time steps in one episode. 
We trained the $K=5$ candidate policies and selected the best policy from these candidates by following the Algorithms \ref{alg:DomainRandomization} and \ref{alg:PolicySelection}. In the Half-Cheetah-Dir domain, the number of tasks was two, forward and backward. Therefore, the size of the context was $|\mathcal{C}|=2$. In the Half-Cheetah-Vel domain, one hundred tasks were randomly sampled from the uniform distribution, thus, the size of the context was $|\mathcal{C}|=100$. In the skill generation phase, we set the maximum number of iterations as $k_\mathrm{max}=100$. Table \ref{tab:HyperParameter} in the Appendix lists the hyperparameters used in our proposed learning framework.
\color{black}

\subsection{Monopod Heading Task}
In the monopod heading task, the robot model tried to hit a ball back to a specified goal point while maintaining its balance until it hit the ball, \textcolor{\hcolor}{as shown in Fig. \ref{fig:snapshots-adapted}(c) (d) (e)}. 
As small changes in the movement result in significantly different reward values, precise dynamic movement control is necessary.
The reward function for the task is defined as follows:
\begin{equation}
    r = \exp \left (- \lambda_1 \| p_\mathrm{ball} - p_\mathrm{targ}(t,c) \|^2 \right )-\lambda_2 \| \tau \|^2, 
\end{equation}
where $p_\mathrm{ball}$ denotes the ball position; $p_\mathrm{targ}(t,c)$ is the context-dependent desired ball trajectory; and $\tau$ denotes the joints torque. The desired ball trajectory was derived using the motion equation from a desired hit position to a target point within a given ball flying duration. The learning trial was terminated if the ball position deviated from the desired ball trajectory by a certain distance. The reward parameters were set to $\lambda_1=10$ and $\lambda_2=5 \times 10^{-3}$. Furthermore, we terminated the learning trial when the distance between the head position and the desired hitting position exceeded a certain threshold, primarily because the monopod robot fell to the ground.

The monopod robot model had five joints, each controlled by a proportional (P) servo controller. The state variables of the robot are listed in Table \ref{tab:Observation}. \textcolor{\hcolor}{Notably, the robot did not explicitly observe the goal position. Instead, the reward signal informed it of the task achievement. The action variables were the target joint angles for the P servo controller at each joint.}
\begin{table}[t]\centering
\caption{Observables of monopod robot model}\label{tab:Observation}
\scriptsize
\begin{tabular}{lrrr}\toprule
    & Dimensionality & Type \\ \midrule
    Center of mass & $2$ & float \\
    Head position & $2$ & float \\
    Foot position & $2$ & float \\
    Joint angle & $5$ & float \\
    Joint angular velocity & $5$ & float \\
    Ball position & $2$ & float \\
    Ball velocity & $2$ & float \\
    One-step before action & $5$& float \\ \midrule
    Total dimensionality & $25$ & \\
\bottomrule
\end{tabular}
\end{table}
In this task, five context variables were considered, which are presented in Table \ref{tab:context parameter}.
The coefficient of restitution defines a physical parameter of the ball and describes the relationship between ball velocities before and after hitting it.
Horizontal and vertical goal position variables define different goal locations, and horizontal and vertical throw-in positions represent different ball throw positions toward the monopod.
\begin{table}[!t]\centering
	\caption{Context variables of monopod robot model}
	\label{tab:context parameter}
	\scriptsize
	\begin{tabular}{lrrl}\toprule
	 &Type & Settings \\\midrule
	Coefficient of restitution &float & \{small, large\}\\
	Horizontal goal position &float & \{near, far\}\\
	Vertical goal position &float & \{low, high\}\\
	Horizontal throw-in position &float & \{near, far\}\\ 
	Vertical throw-in position vertical-axis &float & \{low, hight\}\\ 
	\bottomrule
	\end{tabular}
\end{table}

\subsubsection{\textcolor{\hcolor}{Context randomization}}
\label{sec:policy_generation}
In the foundational policy acquisition phase, we trained the networks under the combination of context variables.
In other words, we conducted context randomization with 32 different combinations of context variables as five context variables with two settings were considered, that is, $2^5=32$.
We repeated this learning process 25 times to generate 25 different domain randomized policies with different random seeds.

\subsubsection{Selecting the foundation policy network}
Using the policy selection method described in Algorithm $2$, only one policy was selected from the 25 candidate policies.

\subsubsection{Evaluating the skill generation performance}
\label{sec:policy_evaluation}
We evaluated how the monopod model can efficiently generate skills to cope with novel tasks and environments using our proposed method.
The parameters representing the novel settings were uniformly sampled from intervals defined in Table \ref{tab:context parameter}.  
Concretely, we randomly sampled the $32$ context and fixed these values to evaluate the task performances of each method.
We confirmed that these $32$ settings differed from those used in the foundational policy acquisition phase described in Section \ref{sec:policy_generation}.
We then compared the proposed approach with the policy acquired using the domain randomization method, which is widely used to cope with an unknown environment.

We then further explored the potential of our proposed approach. We set the goal position in the rear region of the robot, while all training in the foundational policy acquisition phase was done with the goal position in the front region. Thus, we tried to evaluate the extrapolation performance of our skill generation method here. 

\subsubsection{Evaluating policy selection performance}
\label{sec:evaluating_policy_selection}
We compared the proposed policy selection method with random selection in terms of top-one regret, which has been adopted to evaluate policy \mycite{fu2021}.
\textcolor{\hcolor}{The top-one regret evaluates the error between the selected return value of the chosen policy and the best return that acquired the highest return among the candidate policies as follows:}
\begin{align}
    \frac{1}{N}\sum_{i=1}^{N} |R_\mathrm{best} - R_\mathrm{selected}|,
\end{align}
where $R_\mathrm{best}$ denotes the return value of the best policy, and $R_\mathrm{selected}$ denotes that of the selected policy using the proposed method.
To reliably evaluate the top-one regret, we used a bootstrap method.
Concretely, as described in Section \ref{sec:policy_generation}, we generated $25$ candidate policies.
Subsequently, we sampled five policies with replacements from the $25$ candidates.
Through the sampling procedure described above, we randomly made $N=1000$ sets of the five policies.
\section{Results}
\subsection{Multi-locomotion Domains}
\begin{figure}[tbhp]
    \centering
    \begin{minipage}[c]{0.46\linewidth}
        \includegraphics[width=\linewidth]{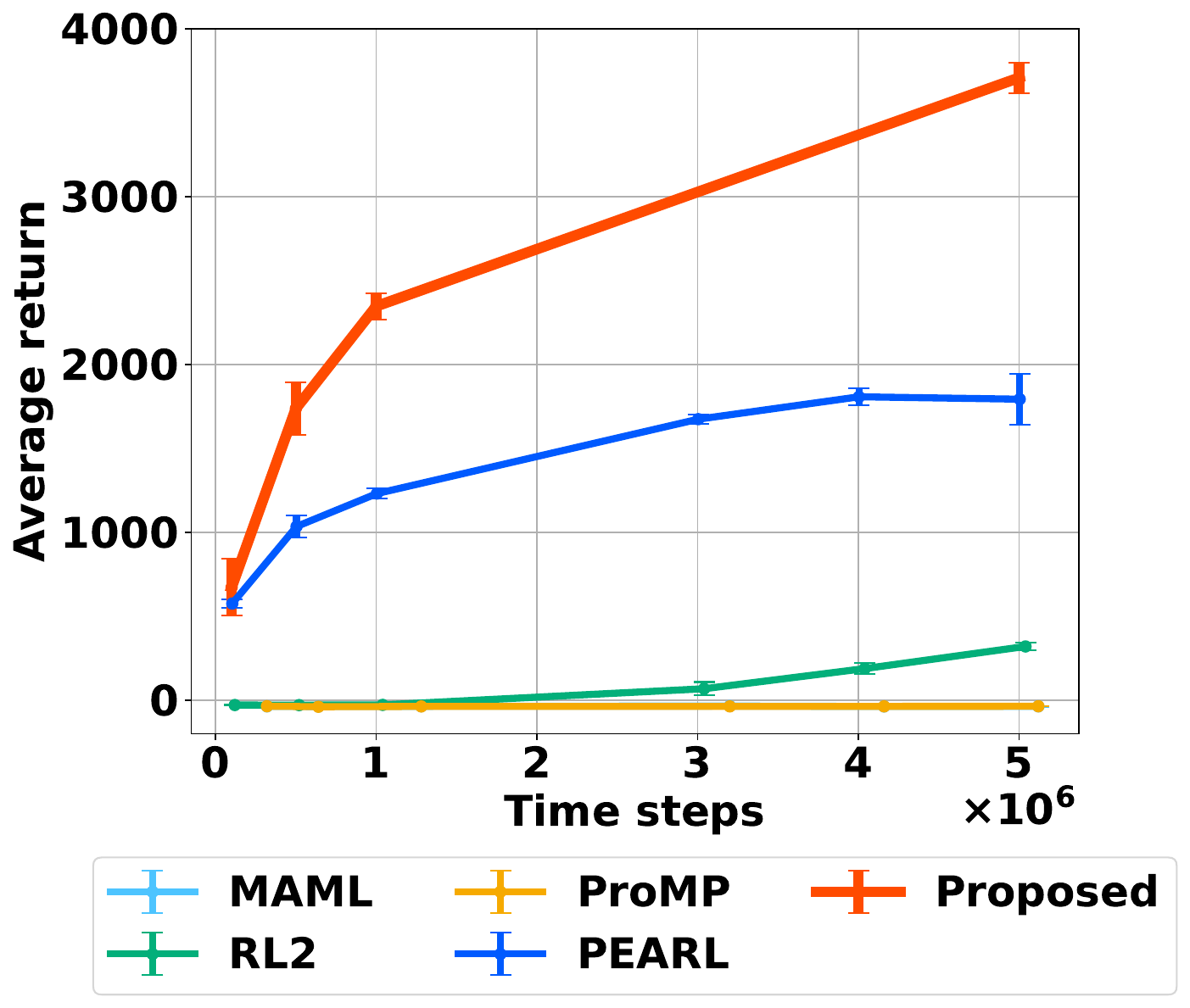}
        \subcaption{Half-Cheetah-Dir}
        \label{fig:half-cheetah-dir-score}
    \end{minipage}
    \begin{minipage}[c]{0.46\linewidth}
        \includegraphics[width=\linewidth]{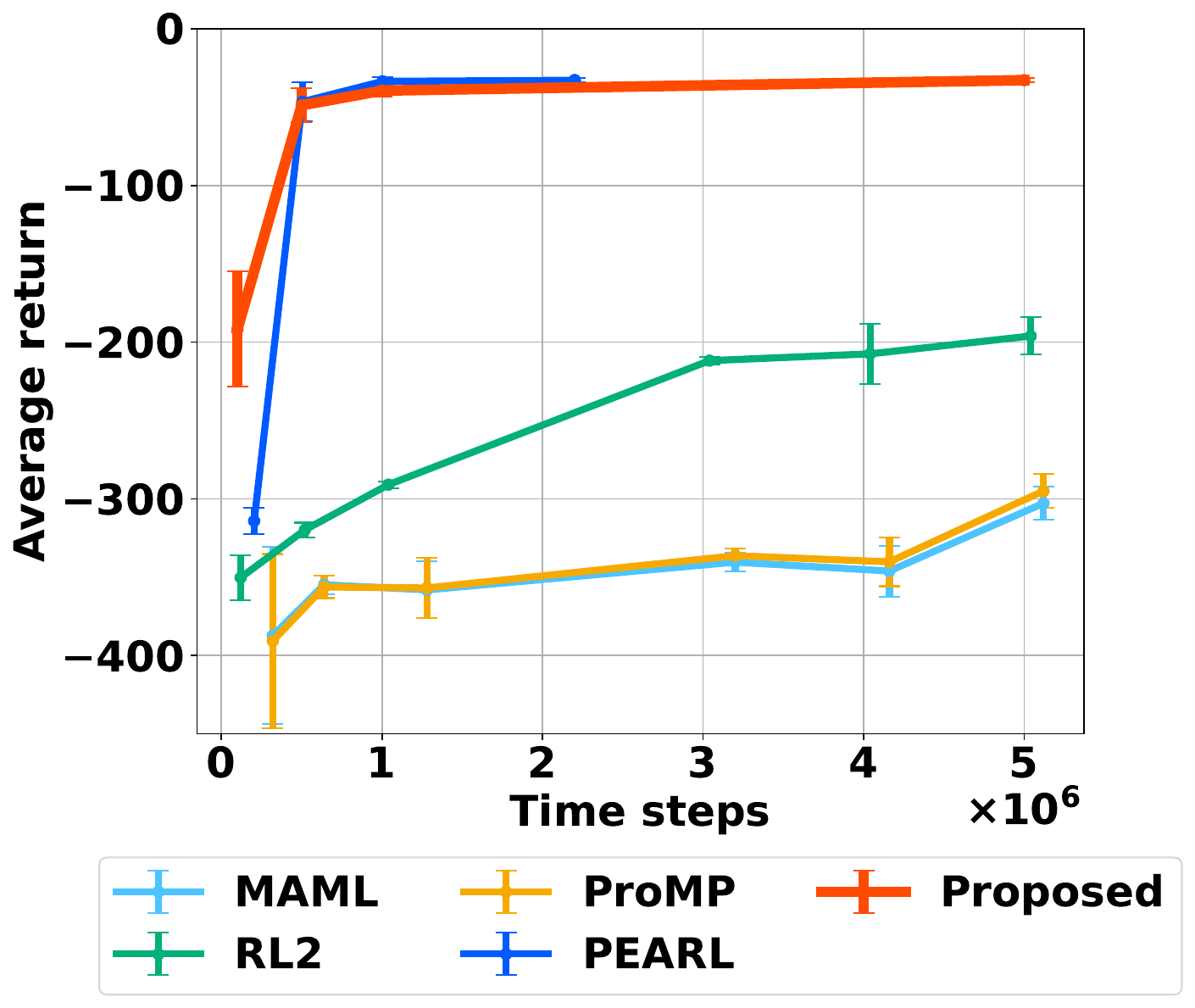}
        \subcaption{Half-Cheetah-Vel}
        \label{fig:half-cheetah-vel-score}            
    \end{minipage}
    \caption{
        \color{\hcolor}
        Average Returns vs. Training Steps. Error bars show the standard deviation.
        (a) Proposed method significantly outperformed all the baseline methods.
        (b) Proposed method showed better learning performance at the early learning stage, and eventually achieved comparable performance with the PEARL method \cite{rakelly2019} and outperformed other three baseline methods.
        Both our proposed and PEARL methods seemed to achieve close to the optimal control performance of the given domain, and the performance of both methods was saturated after about $1$M time steps.
        Proposed method showed comparable performance with the PEARL and outperformed other three baseline methods.
        \color{black}
    }
    \label{fig:half-cheetah-meta-score} 
\end{figure}

The proposed method was evaluated on two Multi-locomotion domains, the Half-Cheetah-Dir domain and the Half-Cheetah-Vel domain, and compared with previous Meta-RL methods, MAML\cite{finn2017}, RL$^2$ w/ PPO\cite{duan2016, rakelly2019}, ProMP \cite{rothfuss2019}, PEARL\cite{rakelly2019}. 
Figure \ref{fig:half-cheetah-meta-score} shows the average returns.
As for our proposed method, the average return was derived over five learning trials.
The training steps correspond to the foundation policy acquisition phase.
For example, since we trained five policies to select one foundation policy, the learning performance at the $500$k time step corresponds to the performance of the foundation policy selected from five candidate policies acquired over $100$k training time steps. 
For the baseline methods, learning performances were extracted from the data provided in \cite{rakelly2019data}.
In the Half-Cheetah-Dir domain, as shown in Fig. \ref{fig:half-cheetah-dir-score}, our proposed method significantly outperformed all the baseline methods. In the Half-Cheetah-Vel domain, as plotted in Fig. \ref{fig:half-cheetah-vel-score}, our proposed method showed much better learning performance at the early learning stage, and eventually achieved comparable performance with the PEARL method \cite{rakelly2019} and outperformed other three baseline methods. Both our proposed and PEARL methods seemed to achieve close to the optimal control performance of the given domain, and the performance of both methods was saturated after about $1$M time steps.

\subsection{Skill Generation Performance}
Domain randomization (DR) is a frequently used approach to cope with multiple tasks and environments.
We first evaluated the acquired monopod heading policies using DR as the baseline.
\textcolor{\hcolor}{We trained the networks under the same combination of context variables as that described in Section \ref{sec:policy_generation}.}
Notably, the proposed method also trained the encoder and policy networks, while the DR method did not.

\begin{figure*}[htbp]
    \centering
    \begin{minipage}[c]{0.65\textwidth}
        \centering
        \begin{minipage}[c]{0.99\textwidth}
            \includegraphics[width=\linewidth]{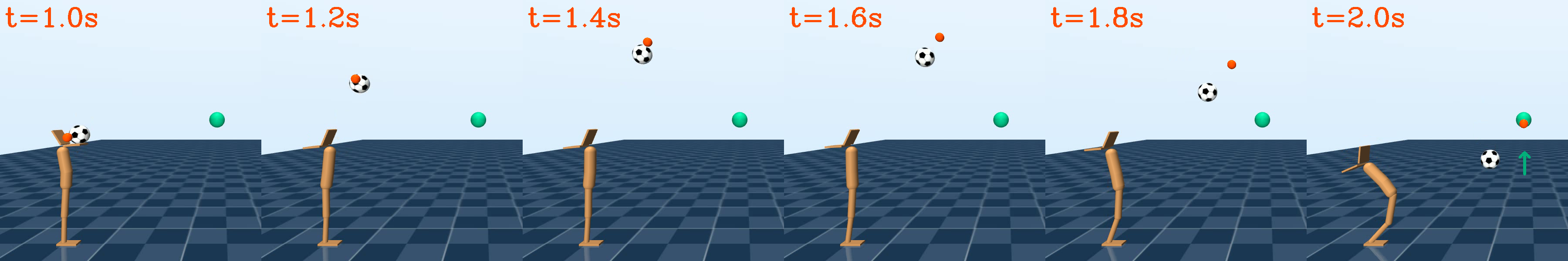}
            \subcaption{Domain randomization}
            \label{fig:DomainRandomizationSnapshot}
        \end{minipage}\\
        \begin{minipage}[c]{0.99\textwidth}
            \includegraphics[width=\linewidth]{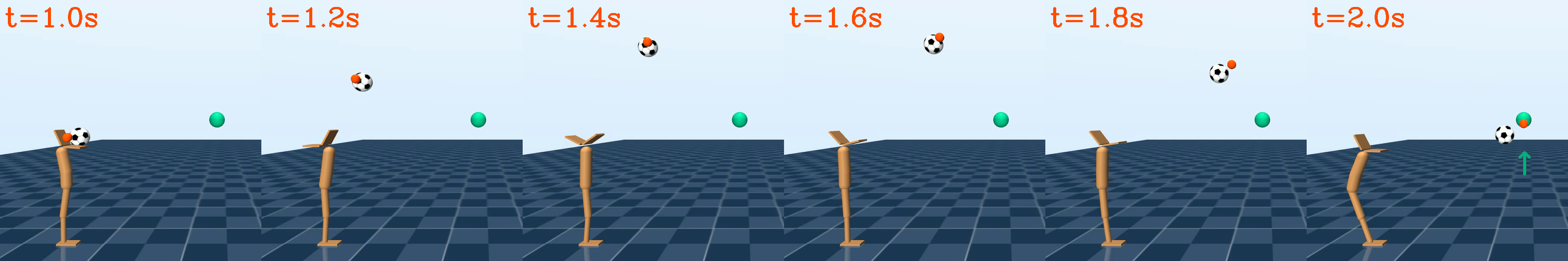}
            \subcaption{Proposed}
            \label{fig:AdaptationSnapshot}            
        \end{minipage}
    \end{minipage}
    \begin{minipage}[c]{0.3\textwidth}
        \includegraphics[width=0.95\linewidth]{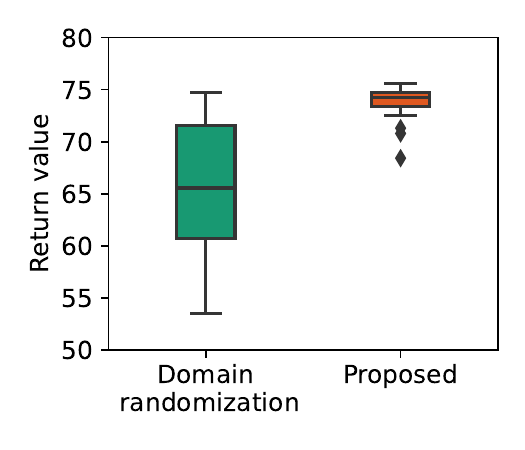}
        \subcaption{Skill generation performance}\label{fig:AdaptationPerformanceBoxPlot}
    \end{minipage}
    \caption{
        Monopod heading performances with the novel goal position.
        Leftmost snapshot shows the timing of hitting the ball, that is, $t=1.0$ s.
        Green sphere indicates the goal position.
        Red dot indicates the desired trajectory.
        (a) Heading performance by domain randomization.
        As shown in the rightmost panel, at $t=2.0$ s, the ball deviated from the goal position.
        (b) Heading performance is performed using the proposed method.
        Hit ball successfully tracked the desired trajectory and passed around the goal position.
        Notably, the monopod also learned to maintain its balance with the hit ball.
        (c) Skill generation performance: domain randomization approach could not achieve a satisfactory multitask performance (green bar).
        Multiple tasks in multiple environments are accomplished using the proposed method (red bar). 
    }
\end{figure*}

Figures. \ref{fig:DomainRandomizationSnapshot} and \ref{fig:AdaptationSnapshot} show the monopod heading task performances at a novel goal position.
As shown in the rightmost panel of Fig. \ref{fig:DomainRandomizationSnapshot}, the ball deviated from the goal position when a policy acquired by DR was adopted.
Using the proposed method, the ball hit by the monopod successfully tracked the desired trajectory and passed around the goal position, as depicted in Fig. \ref{fig:AdaptationSnapshot}.
Notably, the monopod model also learned to maintain balance when hitting the ball.

Figure. \ref{fig:AdaptationPerformanceBoxPlot} shows the task performances of the policies trained by the DR (green bar) and the proposed adaptation (red bar).
Each box plot was depicted based on $32$ task performances, as described in Section \ref{sec:policy_evaluation}.
Both the DR and generation phase of the proposed method trained the networks for $3 \times 10^6$ time steps, which approximately corresponds to $45,000$ trials.
The skill-generation phase only required additional $k_{\mathrm{max}} = 100$ trials to perform novel tasks successfully.

\subsection{Policy Selection Performance}
Here, we evaluate how the selected policy based on the proposed method performed better than a randomly selected model.
We adopted the top-one regret as the performance measure to compare the selection performances.
Section \ref{sec:evaluating_policy_selection} describes a concrete calculation of the top-one regret.

The top-one regret score of the proposed method was $1.4$, whereas that of the random selection was $5.1$.
As the smaller regret showed better performance, we confirmed that the proposed method contributed to selecting a potentially well-performing policy after the adaptation phase. 
\textcolor{\hcolorb}{
In our approach, we used the performance index described in Eq. (\ref{eq:policy-selection-index}) for the policy selection. We selected the best performing policy based on the index. Therefore, we did not need to calibrate any threshold. Furthermore, we used the same hyperparameters of the performance index, $\alpha$, $\beta$, and $\gamma$, for all experimental setups. This suggests that our selection approach was robust to changes in the target tasks.
}

\subsection{Novel Skill Generation Performance}
In this section, we showed how the skill generation phase of our proposed method modifies a policy to cope with a novel task.
We only optimized the one-dimensional latent variable $z$ to improve the task performance for the newly provided task.
The optimization procedure of the latent variable based on the TPE is explained in Section \ref{sec:Adaptation_Phase}.

\begin{figure}[thbp]
    \centering
    \begin{minipage}[b]{0.65\linewidth}
        \begin{minipage}[b]{0.45\linewidth}
            \includegraphics[width=0.99\linewidth]{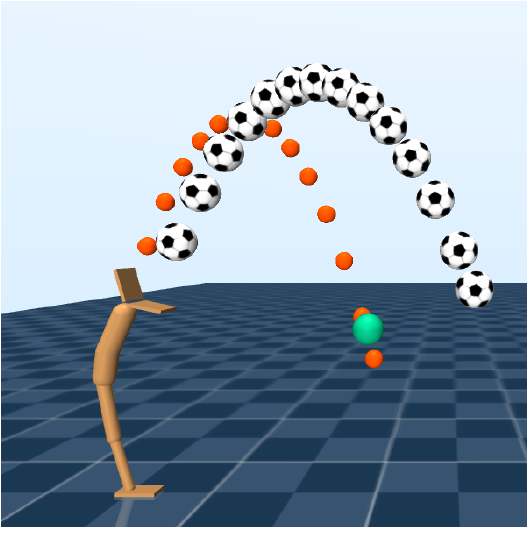}
            \subcaption{Foundational}
            \label{fig:SnapshotBeforeAdapt-before}
        \end{minipage}
        \begin{minipage}[b]{0.45\linewidth}
            \includegraphics[width=0.99\linewidth]{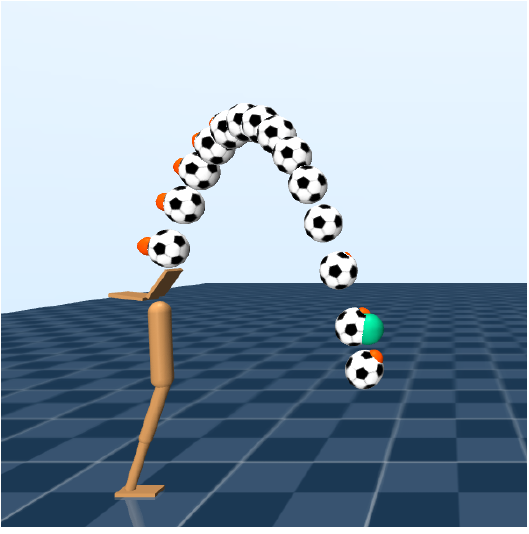}
            \subcaption{Generated skill}
            \label{fig:SnapshotBeforeAfter-after}
        \end{minipage}
    \end{minipage}
    \hspace{-24pt}
    \begin{minipage}[b]{0.33\linewidth}
        \includegraphics[width=1.1\linewidth]{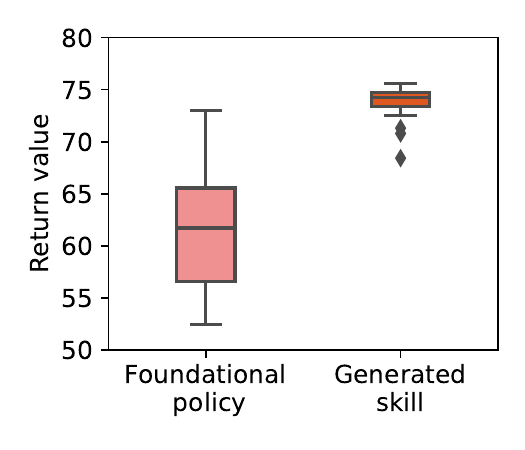}
        \subcaption{Performances}
        \label{fig:BeforeAfterAdaptBarPlot}
    \end{minipage}
    \caption[snapshot before after]{
        Multitask Heading performance on novel tasks.
        Green sphere indicates the goal position.
        Red dot indicates the desired trajectory.
        (a) Foundational policy: the hit ball did not track the desired trajectory and deviated from the goal.
        (b) Generated skill: the adapted policy successfully hit the ball toward the new goal position.
        (c) Multitask heading performances:
        With the foundational policy, task performances were unsatisfactory and varied.
        With the generated skill, generated skills for novel tasks consistently achieved successful heading performances.
    }
    \label{fig:SnapshotBeforeAfter}
\end{figure}

We depicted how the proposed skill generation method improved the task performance in Fig. \ref{fig:SnapshotBeforeAfter}.
As shown in Fig. \ref{fig:SnapshotBeforeAdapt-before}, the hit ball did not track the desired trajectory and deviated from the goal before the adaptation.
To evaluate task performance before the skill generation, we set the latent variable $z=0$, assuming that we have prior knowledge of the novel task.
After the skill generation, the optimized skill successfully hit the ball towards the new goal position, as shown in Fig. \ref{fig:SnapshotBeforeAfter-after}.

We also evaluated the overall performances of the policies before and after the skill generation on $32$ different novel tasks, wherein the tasks were sampled randomly, as described in Section \ref{sec:policy_evaluation}.
Figure \ref{fig:BeforeAfterAdaptBarPlot} shows the task performances of each policy.
Before the skill generation, for most novel tasks, task performances were unsatisfactory.
Furthermore, performances were varied.
After the skill generation, the generated policy for novel tasks consistently achieved successful heading performances.

\begin{figure*}[thbp]
    \begin{minipage}[c]{0.99\textwidth}
        \includegraphics[width=\linewidth]{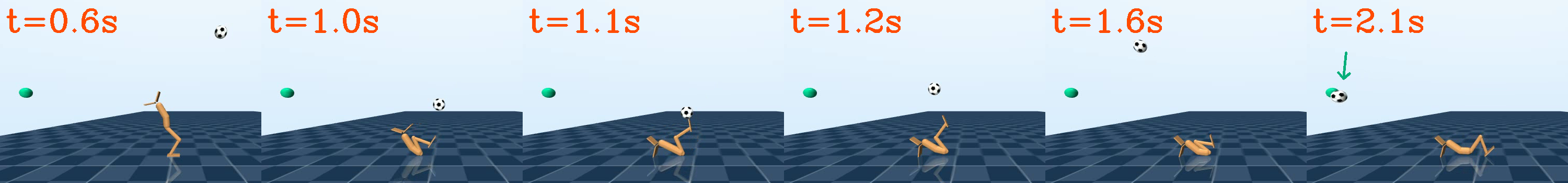}
    \end{minipage}
    \caption{
        Generated overhead kicking skill. We set the goal position (green) in the rear region of the robot, while all training in the foundational policy acquisition phase was done with the goal position in the front region. Thus, here we tried to evaluate the extrapolation performance of our skill generation method. As a result, the robot model generated a whole new motor skill to ``kick'' the ball backward to the goal position instead of heading it.
    }
    \label{fig:BackkickSnapshot}     
\end{figure*}

Furthermore, when we placed the goal position at the back of the robot, the robot model generated a whole new motor skill to ``kick'' the ball backward to the goal position instead of heading it.
Figure \ref{fig:BackkickSnapshot} illustrated the generated overhead kick movement to hit the ball to the totally new goal position.

The above results indicate that the proposed low-dimensional skill representation and derivative-free optimization successfully facilitate skill emergence in novel situations.

\begin{figure}[thbp]
    \begin{minipage}{0.45\linewidth}
        \centering
        \includegraphics[width=0.95\linewidth]{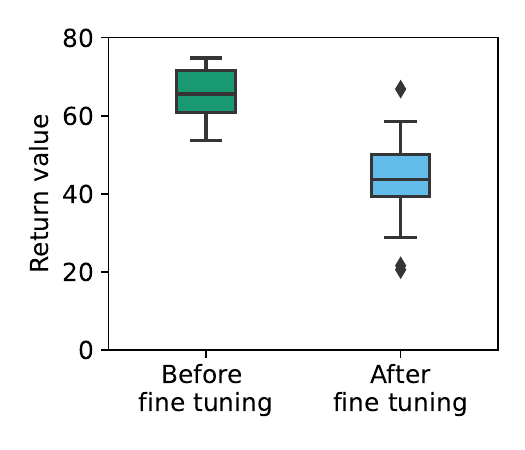}
        \subcaption{Fine-tuning performance}
        \label{fig:FineTuningBoxPlot}
    \end{minipage}
    \begin{minipage}{0.45\linewidth}
        \includegraphics[width=0.95\linewidth]{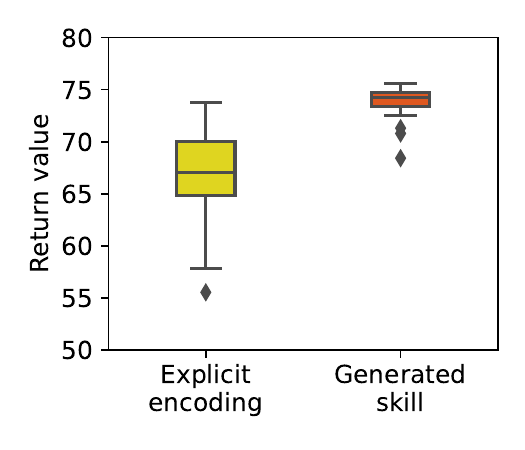}
        \subcaption{Skill generation performance}
        \label{fig:EncoderBoxPlot}
    \end{minipage}
    \caption[Fine-tuning box plot]{
        (a) Fine-tuning performance.
        We found that task performances became worse than those before fine-tuning trials.
        This result indicated that the policy model could not adapt to new domains with only a small number of additional trials.
        (b) Proposed method was compared with the baseline method using context variable information.
        Interestingly, the proposed skill-generation method showed a better heading performance than the method using context information.
        This result empirically showed that we extracted a meaningful latent space for the given movement category.
    }    
\end{figure}

\section{Discussion}
\subsection{Skill Generation through Latent Variable Optimization}
One may consider that the comparison presented in Fig. \ref{fig:AdaptationPerformanceBoxPlot} is not fair because the proposed method conducted additional $100$ trials for adaptation.
Therefore, we conducted fine-tuning trials on the novel domain to observe how the policy trained by domain randomization can improve with additional $100$ learning iterations.

Figure \ref{fig:FineTuningBoxPlot} shows the heading performances with the $32$ randomly selected domains.
We found that the task performances became worse than those before the fine-tuning trials.
This result indicated that the domain randomization policy model could not adapt to new domains with only a small number of additional trials.
In other words, extracting the low-dimensional latent space and optimizing the latent variable were significant procedures for generating a skill to cope with novel tasks and environments.

\subsection{Using Explicit Context Information}
In this study, we assumed that the policy could not observe the context variable in the adaptation phase as we usually lack information regarding novel domains. 
However, it would be interesting to investigate how task performance can be improved if the context variable was observable.
Here, we explicitly encoded the context variable of $32$ novel domains using the trained encoder in the skill-generation phase.
The encoder forwards the latent variable to the policy network.

Figure \ref{fig:EncoderBoxPlot} shows the task performances using explicit context information.
Interestingly, the proposed adaptation method showed better heading performances than the method using context information.
Since the explicit encoder has not been trained in the context of novel domains, the performance degrades. However, the generation skill can adapt to the novel context using the reward signal. Consequently, our proposed method achieved better performance than the method that considers the context variables as inputs.
This result empirically showed that we extracted a meaningful latent space for the given movement category.  

\subsection{Limitations}
Our three-phase multitask RL method requires the foundational policy acquisition phase in which multiple candidate policies must be trained. However, the number of required learning trials grows only linearly with the number of candidate policies. Therefore, we have shown that our proposed method exhibits better learning performance than the baseline methods even when explicitly considering the number of learning trials required to train the multiple candidate policies. On the other hand, the number of required candidate policies is still a hyperparameter for our learning system. Although we can parallelize the computation of candidate policy learning, as a future study, we are interested in developing a method to regularize the learning process to guide an acquired policy to be adaptable to generate skills for novel tasks, rather than generating multiple candidate policies and selecting one among the candidates, to further reduce the computational complexity of the learning framework. 
\textcolor{\hcolorb}{
In this study, we have shown that acquisition of the foundation policy is the key to adaptation to new task goals and environments. However, adaptation to completely different task domains has not been solved even for the large-scale model trained with large-scale data \mycite{Kalashnikov2021, vuong2023}. As a future study, we will investigate whether our proposed approach can deal with a wider range of domains by considering a wider range of contexts in the foundation policy learning with the use of parallel computation in multiple candidate policy learning processes.
}

\section{Conclusion}
\textcolor{\hcolor}{
In motor skill learning, the acquisition of policies that allow rapid adaptation to new situations with different environments and tasks is an important issue. Through the standard multi-locomotion benchmarks and the heading task using the monopod robot, it was verified that the proposed three-phase learning method can quickly generate policies that can cope with multiple tasks. Our framework for learning a foundational policy model based on encoder learning, which embeds explicit contexts in latent spaces, and for selecting and generating motor skills, can efficiently embed and extract the ability to handle multiple tasks. Our proposed method outperformed baseline multitask RL methods on the multi-locomotion benchmarks. Furthermore, by exploiting this skill generation ability in the monopod heading task, we have successfully generated an overhead kicking skill that was not included in the foundational policy acquisition phase at all.}

As a future study, we aim to explore the possibility of the proposed method on a multi-agent learning problem.
We are interested in observing how our proposed method can modify agent policy according to implicit changes in other agent policies.

\section*{Acknowledgments}
This work was supported by JSPS KAKENHI (Grant number: JP19J22987, JP22H04998, JP23K24925); by project JPNP20006, commissioned by NEDO; by JST Moonshot R\&D program (Grant number: JPMJMS223B-3); and by Tateisi Science and Technology Foundation.

{
\section*{Appendix}
\subsection*{Learning Algorithm}
We adopted the target networks, clipped double-$Q$ learning, and entropy autotuning in the foundational policy acquisition.

\subsubsection*{Clipped double-$Q$ learning}
double $Q$-function networks were trained to reduce Bellman residuals, as shown in Eq. (\ref{eq:ClippedDQ}) as follows:
\begin{align}
    &\min_{\phi_i} \mathbb{E}_{\mathcal{D}}
        \left ( Q(s, a, c; \phi_i) - r(s, a, c) - \gamma \bar{V}(s', c) \right )^2,
    \label{eq:ClippedDQ} \\
    &\bar{V}(s, c) := \min_{i\in {1, 2}} Q(s, a', c;\phi_i^\mathrm{target}) - \alpha \log \pi(a'|s, z;\psi), \nonumber\\
    & a' \sim \pi(a'|s, z;\psi), z\sim q(z|c;\eta), \{s, a, s', c\}\sim \mathcal{D}, \nonumber
\end{align}
where $\phi_i$ and $\phi_{\mathrm{target}}$ are the $Q$-network parameters;
$r$, $\gamma$, and $\alpha$ denote the reward function, discount factor, and temperature parameter, respectively;
$\pi$ represents the policy;
$q$ represents the encoder;
and $\mathcal{D}$ is the state-action pair data stored in a replay buffer.
The value function $\bar{V}$ was computed from $Q(s', a', c)$, where $s'$, $a'$ and $c$ are the state, action, and context at the next step, respectively. 
Simply using the above method tends to overestimate the action values;
therefor, clipped double-$Q$ Learning was proposed to overcome this overestimation problem \mycite{fujimoto2018}.
To use this method, we trained two networks and selected the smaller output of the two approximated $Q$ functions to derive the value $\bar{V}$.
% In practice, $Q$ values at the next step are derived from the target network $\phi_\mathrm{target}$ explained below.

\subsubsection*{Target networks}
The target networks $\phi_\mathrm{target}$ were used to stabilize the learning process of the $Q$ function.
As we explained in the previous section, the target value $V$ was derived from the $Q$ network, which caused the target value estimation to be unstable \mycite{mnih2015human}.
Therefore, as suggested in \mycite{lillicrap2016}, we updated the target network through moving average as follows:
\begin{align}
    \phi_i^\mathrm{target} \leftarrow \tau \phi_i + (1 - \tau) \phi_i^\mathrm{target}.
    \label{eq:TargetNetwork}
\end{align}
The Polyak's coefficient $\tau$, which ranges from 0 to 1, was used to adjust the update rate.

\subsubsection*{Entropy autotuning}
To autotune the temperature parameter $\alpha$ of the entropy regularization term, we considered the following optimization problem \mycite{haarnoja2018}: 
$H$ is the conditional entropy of the policy distribution given the state $s$, and $\bar{H}$ is the target entropy set by an experimenter.
\begin{align}
    \max_\pi &\ \mathbb{E}_{\tau \sim \mathcal{D}} 
        \left [ \sum_t \gamma^t r_t \right ], \\
    \mathrm{s.t.} &\ \mathbb{E}_{s \sim \mathcal{D}}[H(\pi)] < \bar{H}.
    \label{eq:MDPEntropyUppderBound}
\end{align}
Although this equation does not explicitly contain the temperature parameter $\alpha$,
it can be attributed to the following min-max problem as Lagrange duality when the temperature parameter $\alpha$ is regarded as a Lagrange multiplier:
\begin{align}
    \min_{\alpha>0}\max_\pi \mathbb{E}_{\tau\sim \mathcal{D}}  
        &\left [  \sum_t \gamma^t r_t -\alpha (\log \pi_t + \bar{H})\right ].
    \label{eq:LaugrangeDualQ}
\end{align}
The min-max problem was solved as:
\begin{align}
    \min_{\alpha>0} \mathbb{E}_{(a,s,z) \sim \mathcal{D}}  
        &\left [ \alpha (-\log \pi(a|s, z) - \bar{H})\right ].
    \label{eq:ObjectFuncAlpha}
\end{align}
The temperature parameter $\alpha$ is transformed $\alpha \leftarrow \exp(\alpha)$ to guarantee the positive value as practical implementation
\footnote{https://github.com/rail-berkeley/softlearning/blob/master/softlearning/\\algorithms/sac.py\#L140. Accessed: July-10, 2023.}.

\subsection*{Policy Gradient and Reparametrization Trick}
We used the reparametrization trick to calculate the gradient of the objective function with respect to policy parameters.
The objective function was expressed as follows:
\begin{align}
&\max_{\psi, \eta}  \mathbb{E}_{(s, c)\sim \mathcal{D}} [\bar{V}(s, c)]  - \beta \mathbb{E}_{c\sim \mathcal{D}} \mathrm{KL}\left[ q \| \rho \right ], \\
& a' \sim \pi(a'|s, z;\psi), z\sim q(z|c;\eta), \rho(z)=\mathcal{N}(z|\mathbf{0},\sigma \mathbf{I}). \nonumber
\end{align}
Here, $\phi$ and $\eta$ are the network parameters of the policy $\pi$ and encoder $q$ networks, respectively; and
$\rho(z)$ is the prior distribution of the latent variable.

To derive the gradient, expectation over the policy distribution $\pi$ and encoder distribution $q$ must be evaluated as follows:
\begin{alignat}{2}
&\max_{\psi, \eta}&\ &  \mathbb{E}_{(s,c)\sim\mathcal{D}, q\pi} [Q(s,a,c) - \alpha \log \pi (a|s, z;\psi)]  \nonumber\\
& && - \beta \mathbb{E}_{c\sim \mathcal{D}, q} \left[ \log \frac{q(z|c; \eta)}{\rho(z)} \right ].
\end{alignat}
By defining the related variables, as shown below, the parameters of the neural network do not depend directly on the distributions.
\begin{align}
a &= \tanh(f(s,z; \psi) + \sigma(s, z; \psi) \epsilon), \epsilon \sim \mathcal{N}(\mathbf{0}, \mathbf{I}), \\
z &= g(c; \eta) + \delta\omega, \omega \sim \mathcal{N}(\mathbf{0}, \mathbf{I}),\\
&P(a;\psi) = P(\epsilon) \left| \frac{da}{d\epsilon}\right |^{-1}_\psi, P(z;\eta) = P(\omega) \left| \frac{dz}{d\omega}\right |^{-1}_\eta.
\end{align}
Here, $\epsilon$ and $\omega$ are the random variables sampled from the standard normal distribution; and
$f$, $\sigma$, and $g$ are the functions represented by the neural networks.
Now, the objective function can be defined as,
\begin{alignat}{1}
J(\psi, \eta)=&\  \mathbb{E}_{(s,c)\sim\mathcal{D}, \omega, \epsilon} [Q(s,a,c) - \alpha \log \pi (a|s, z;\psi)]  \nonumber\\
& - \beta \mathbb{E}_{c\sim \mathcal{D}, \omega} \left[ \log \frac{q(z|c;\eta)}{\rho(z)} \right ], \nonumber \\
& a = a(s, \epsilon, z; \psi), z = z(c, \omega; \eta), \nonumber
\end{alignat}
where the integrated variables, $\epsilon$ and $\omega$, are independent of the network parameters.
Therefore, the interchange of the gradient and integral for the policy and encoder parameters, $\psi$ and $\eta$, was allowed:
\begin{align}
\nabla J &= \nabla \mathbb{E}_{(s,c)\sim\mathcal{D}, \omega, \epsilon}\left [ Q - \alpha \log \pi - \beta \log \frac{q}{\rho} \right ], \nonumber\\
&= \mathbb{E}_{(s,c)\sim\mathcal{D}, \omega, \epsilon} \left [ \nabla Q - \alpha \nabla \log \pi - \beta \nabla \log \frac{q}{\rho} \right ].
\end{align}

\subsection*{Hyperparameters}
The parameters used in this study are listed in Table \ref{tab:HyperParameter}.
Notably, most parameters are based on the original SAC \mycite{haarnoja2018}.

\begin{table}[!htpb]
\centering
\caption{List of hyperparameters}\label{tab:HyperParameter}
\scriptsize
\begin{tabular}{lrrrr}\toprule
hyperparameter &policy search  &adaptation & info \\\midrule
action time period & 50 / 25 ms& 50 / 25 ms & Cheetah / Monopod \\
Q learning rate &0.001 & - \\
$\beta$ &$e^{-5}$ &$e^{-5}$ \\
replay buffer size &$10^6$ &100 \\
collect step size per training &1000 &200/100 &  Cheetah / Monopod\\
collect step size per evaluation &200/1000 & 200/100 &  Cheetah / Monopod\\
update times per training & 1000 & 10\\
batch size &256 &100 \\
discount $\gamma$ &0.99 &1.00 \\
optimization algorithm &Adam & TPE \\
encoder variance $\delta$ &0.01 &0.01 \\
prior variance $\sigma$ &1 &1 \\
temperature $\alpha$ & -& 0.01 \\
target entropy $\bar{H}$ & $-|A|$& - \\
number of candidates & 5 / 25 & -& Cheetah / Monopod\\
skill generation iterations $k_\mathrm{max}$ &-&  100\\
\bottomrule
\end{tabular}
\end{table}
}
\bibliographystyle{IEEEtran}
\bibliography{no-need-url, need_url_item}

% Generated by IEEEtran.bst, version: 1.14 (2015/08/26)
\begin{thebibliography}{10}
\providecommand{\url}[1]{#1}
\csname url@samestyle\endcsname
\providecommand{\newblock}{\relax}
\providecommand{\bibinfo}[2]{#2}
\providecommand{\BIBentrySTDinterwordspacing}{\spaceskip=0pt\relax}
\providecommand{\BIBentryALTinterwordstretchfactor}{4}
\providecommand{\BIBentryALTinterwordspacing}{\spaceskip=\fontdimen2\font plus
\BIBentryALTinterwordstretchfactor\fontdimen3\font minus \fontdimen4\font\relax}
\providecommand{\BIBforeignlanguage}[2]{{%
\expandafter\ifx\csname l@#1\endcsname\relax
\typeout{** WARNING: IEEEtran.bst: No hyphenation pattern has been}%
\typeout{** loaded for the language `#1'. Using the pattern for}%
\typeout{** the default language instead.}%
\else
\language=\csname l@#1\endcsname
\fi
#2}}
\providecommand{\BIBdecl}{\relax}
\BIBdecl

\bibitem{clavera2019}
A.~Nagabandi, I.~Clavera, S.~Liu, R.~S. Fearing, P.~Abbeel, S.~Levine \emph{et~al.}, ``Learning to adapt in dynamic, real-world environments through meta-reinforcement learning,'' in \emph{7th {{International Conference}} on {{Learning Representations}}}.\hskip 1em plus 0.5em minus 0.4em\relax {New Orleans, LA, USA}: {OpenReview.net}, 2019.

\bibitem{rakelly2019}
K.~Rakelly, A.~Zhou, C.~Finn, S.~Levine, and D.~Quillen, ``Efficient off-policy meta-reinforcement learning via probabilistic context variables,'' in \emph{Proceedings of the 36th {{International Conference}} on {{Machine Learning}}}, K.~Chaudhuri and R.~Salakhutdinov, Eds., vol.~97.\hskip 1em plus 0.5em minus 0.4em\relax {Long Beach, California, USA}: {PMLR}, 2019, pp. 5331--5340.

\bibitem{chen2022}
J.~Chen and H.~Qiao, ``Motor-cortex-like recurrent neural network and multitask learning for the control of musculoskeletal systems,'' \emph{{IEEE} Transactions on Cognitive and Developmental Systems}, vol.~14, no.~2, pp. 424--436, 2022.

\bibitem{jin2024}
C.~Jin, X.~Feng, and H.~Yu, ``A brain-inspired incremental multitask reinforcement learning approach,'' \emph{{IEEE} Transactions on Cognitive and Developmental Systems}, vol.~16, no.~3, pp. 1147--1160, 2024.

\bibitem{Krakauer2019}
J.~W. Krakauer, A.~M. Hadjiosif, J.~Xu, A.~L. Wong, and A.~M. Haith, \emph{Motor Learning}.\hskip 1em plus 0.5em minus 0.4em\relax John Wiley \& Sons, Ltd, 2019, pp. 613--663.

\bibitem{Morehead2021}
J.~R. Morehead and J.~O. de~Xivry, ``A synthesis of the many errors and learning processes of visuomotor adaptation,'' \emph{bioRxiv 2021.03.14.435278}, 2021.

\bibitem{Tsay2022}
J.~S. Tsay, H.~Kim, A.~M. Haith, and R.~B. Ivry, ``Understanding implicit sensorimotor adaptation as a process of proprioceptive re-alignment,'' \emph{eLife}, vol.~11, p. e76639, aug 2022.

\bibitem{finn2017}
C.~Finn, P.~Abbeel, and S.~Levine, ``Model-agnostic meta-learning for fast adaptation of deep networks,'' in \emph{Proceedings of the 34th {{International Conference}} on {{Machine Learning}}}, D.~Precup and Y.~W. Teh, Eds., vol.~70.\hskip 1em plus 0.5em minus 0.4em\relax {Sydney, Australia}: {PMLR}, 2017, pp. 1126--1135.

\bibitem{andrychowicz2017}
M.~Andrychowicz, F.~Wolski, A.~Ray, J.~Schneider, R.~Fong, P.~Welinder \emph{et~al.}, ``Hindsight experience replay,'' in \emph{Advances in {{Neural Information Processing Systems}}}, I.~Guyon, U.~V. Luxburg, S.~Bengio, H.~Wallach, R.~Fergus, S.~Vishwanathan \emph{et~al.}, Eds., vol.~30.\hskip 1em plus 0.5em minus 0.4em\relax {Long Beach, CA, USA}: {Curran Associates, Inc.}, 2017, pp. 5048--5058.

\bibitem{schaul2015}
T.~Schaul, D.~Horgan, K.~Gregor, and D.~Silver, ``Universal value function approximators,'' in \emph{Proceedings of the 32nd International Conference on Machine Learning}, F.~Bach and D.~Blei, Eds., vol.~37.\hskip 1em plus 0.5em minus 0.4em\relax {Lille, France}: {PMLR}, 2015, pp. 1312--1320.

\bibitem{liu2022}
M.~Liu, M.~Zhu, and W.~Zhang, ``Goal-conditioned reinforcement learning: Problems and solutions,'' in \emph{Proceedings of the Thirty-First International Joint Conference on Artificial Intelligence}, L.~D. Raedt, Ed.\hskip 1em plus 0.5em minus 0.4em\relax {Vienna, Austria}: {ijcai.org}, 2022, pp. 5502--5511.

\bibitem{whitehead1993}
S.~Whitehead, J.~Karlsson, and J.~Tenenberg, ``Learning multiple goal behavior via task decomposition and dynamic policy merging,'' in \emph{Robot Learning. {{The Springer International Series}} in {{Engineering}} and {{Computer Science}}}, J.~H. Connell and S.~Mahadevan, Eds.\hskip 1em plus 0.5em minus 0.4em\relax {Boston, MA, USA}: {Springer US}, 1993, vol. 233, pp. 45--78.

\bibitem{singh1991}
S.~P. Singh, ``Transfer of learning across compositions of sequential tasks,'' in \emph{Proceedings of the {{Eighth International Workshop}} ({{ML91}})}, L.~A. Birnbaum and G.~C. Collins, Eds.\hskip 1em plus 0.5em minus 0.4em\relax {Evanston, Illinois, USA}: {Morgan Kaufmann}, 1991, pp. 348--352.

\bibitem{ude2010}
A.~Ude, A.~Gams, T.~Asfour, and J.~Morimoto, ``Task-specific generalization of discrete and periodic dynamic movement primitives,'' \emph{{IEEE} Transactions Robotics}, vol.~26, no.~5, pp. 800--815, 2010.

\bibitem{wu2022}
H.~Wu, W.~Yan, Z.~Xu, and X.~Zhou, ``A framework of improving human demonstration efficiency for goal-directed robot skill learning,'' \emph{{IEEE} Transactions on Cognitive and Developmental Systems}, vol.~14, no.~4, pp. 1743--1754, 2022.

\bibitem{nair30}
A.~Nair, S.~Bahl, A.~Khazatsky, V.~Pong, G.~Berseth, and S.~Levine, ``Contextual imagined goals for self-supervised robotic learning,'' in \emph{Proceedings of the 3rd {{Annual Conference}} on {{Robot Learning}}}, L.~P. Kaelbling, D.~Kragic, and K.~Sugiura, Eds., vol. 100.\hskip 1em plus 0.5em minus 0.4em\relax {Osaka, Japan}: {PMLR}, 2019, pp. 530--539.

\bibitem{foster2002}
D.~J. Foster and P.~Dayan, ``Structure in the space of value functions,'' \emph{Machine Learning}, vol.~49, no. 2-3, pp. 325--346, 2002.

\bibitem{gulcehre2020}
C.~Gulcehre, Z.~Wang, A.~Novikov, T.~Paine, S.~G{\'o}mez, K.~Zolna \emph{et~al.}, ``{{RL Unplugged}}: A suite of benchmarks for offline reinforcement learning,'' in \emph{Advances in {{Neural Information Processing Systems}}}, H.~Larochelle, M.~Ranzato, R.~Hadsell, M.~F. Balcan, and H.~Lin, Eds., vol.~33.\hskip 1em plus 0.5em minus 0.4em\relax {virtual}: {Curran Associates, Inc.}, 2020, pp. 7248--7259.

\bibitem{slotine1991applied}
J.~E. Slotine and W.~Li, \emph{Applied Nonlinear Control}.\hskip 1em plus 0.5em minus 0.4em\relax {Englewood Cliffs, New Jersey, USA}: {Prentice Hall}, 1991.

\bibitem{morimoto2002}
J.~Morimoto and C.~G. Atkeson, ``Minimax differential dynamic programming: An application to robust biped walking,'' in \emph{Advances in {{Neural Information Processing Systems}}}, S.~Becker, S.~Thrun, and K.~Obermayer, Eds., vol.~15.\hskip 1em plus 0.5em minus 0.4em\relax {Vancouver, British Columbia, Canada}: {MIT Press}, 2002, pp. 1563--1570.

\bibitem{morimoto2009}
J.~Morimoto and C.~G.~Atkeson, ``Nonparametric representation of an approximated {{Poincar\'e}} map for learning biped locomotion,'' \emph{Autonomous Robots}, vol.~27, no.~2, pp. 131--144, 2009.

\bibitem{yu2017}
W.~Yu, J.~Tan, C.~K. Liu, and G.~Turk, ``Preparing for the unknown: Learning a universal policy with online system identification,'' in \emph{Proceedings of {{Robotics}}: {{Science}} and {{Systems}} {XIII}}, N.~M. Amato, S.~S. Srinivasa, N.~Ayanian, and S.~Kuindersma, Eds., {Cambridge, Massachusetts, USA}, 2017.

\bibitem{yu2019a}
W.~Yu, V.~C.~V. Kumar, G.~Turk, and C.~K. Liu, ``Sim-to-real transfer for biped locomotion,'' in \emph{2019 {{IEEE}}/{{RSJ International Conference}} on {{Intelligent Robots}} and {{Systems}} ({{IROS}})}.\hskip 1em plus 0.5em minus 0.4em\relax {Macau, China}: {IEEE}, 2019, pp. 3503--3510.

\bibitem{peng2020}
X.~B. Peng, E.~Coumans, T.~Zhang, T.~E. Lee, J.~Tan, and S.~Levine, ``Learning agile robotic locomotion skills by imitating animals,'' in \emph{Proceedings of {{Robotics}}: {{Science}} and {{Systems XVI}}}, {Held Virtually}, 2020.

\bibitem{wierstra2014}
D.~Wierstra, T.~Schaul, T.~Glasmachers, Y.~Sun, J.~Peters, and J.~Schmidhuber, ``Natural evolution strategies,'' \emph{The Journal of Machine Learning Research}, vol.~15, no.~27, pp. 949--980, 2014.

\bibitem{salimans2017}
\BIBentryALTinterwordspacing
T.~Salimans, J.~Ho, X.~Chen, S.~Sidor, and I.~Sutskever, ``Evolution strategies as a scalable alternative to reinforcement learning,'' \emph{arXiv:1703.03864 [stat.ML]}, 2017. [Online]. Available: \url{https://arxiv.org/abs/1703.03864}
\BIBentrySTDinterwordspacing

\bibitem{bergstra2011}
J.~Bergstra, R.~Bardenet, Y.~Bengio, and B.~K{\'e}gl, ``Algorithms for hyper-parameter optimization,'' in \emph{Advances in {{Neural Information Processing Systems}}}, J.~{Shawe-Taylor}, R.~Zemel, P.~Bartlett, F.~Pereira, and K.~Q. Weinberger, Eds., vol.~24.\hskip 1em plus 0.5em minus 0.4em\relax {Granada, Spain}: {Curran Associates, Inc.}, 2011, pp. 2546--2554.

\bibitem{bergstra2013}
J.~Bergstra, D.~Yamins, and D.~Cox, ``Making a science of model search: Hyperparameter optimization in hundreds of dimensions for vision architectures,'' in \emph{Proceedings of the 30th {{International Conference}} on {{Machine Learning}}}, ser. {{PMLR}}, S.~Dasgupta and D.~McAllester, Eds., vol.~28, no.~1.\hskip 1em plus 0.5em minus 0.4em\relax {Atlanta, Georgia, USA}: {PMLR}, 2013, pp. 115--123.

\bibitem{haarnoja2018a}
T.~Haarnoja, A.~Zhou, P.~Abbeel, and S.~Levine, ``Soft actor-critic: Off-policy maximum entropy deep reinforcement learning with a stochastic actor,'' in \emph{Proceedings of the 35th {{International Conference}} on {{Machine Learning}}}, J.~Dy and A.~Krause, Eds., vol.~80.\hskip 1em plus 0.5em minus 0.4em\relax {Stockholm, Sweden}: {PMLR}, 2018, pp. 1861--1870.

\bibitem{levine2018}
\BIBentryALTinterwordspacing
S.~Levine, ``Reinforcement learning and control as probabilistic inference: Tutorial and review,'' \emph{arXiv:1805.00909 [cs.LG]}, 2018. [Online]. Available: \url{http://arxiv.org/abs/1805.00909}
\BIBentrySTDinterwordspacing

\bibitem{akiba2019}
T.~Akiba, S.~Sano, T.~Yanase, T.~Ohta, and M.~Koyama, ``Optuna: A next-generation hyperparameter optimization framework,'' in \emph{Proceedings of the 25th {{ACM SIGKDD}} International Conference on Knowledge Discovery \& Data Mining}, A.~Teredesai, V.~Kumar, Y.~Li, R.~Rosales, E.~Terzi, and G.~Karypis, Eds.\hskip 1em plus 0.5em minus 0.4em\relax {Anchorage, AK, USA}: {ACM}, 2019, pp. 2623--2631.

\bibitem{todorov2012}
E.~Todorov, T.~Erez, and Y.~Tassa, ``{{MuJoCo}}: A physics engine for model-based control,'' in \emph{2012 {{IEEE}}/{{RSJ International Conference}} on {{Intelligent Robots}} and {{Systems}} ({{IROS}})}.\hskip 1em plus 0.5em minus 0.4em\relax {Vilamoura, Portugal}: {IEEE}, 2012, pp. 5026--5033.

\bibitem{fu2021}
J.~Fu, M.~Norouzi, O.~Nachum, G.~Tucker, Z.~Wang, A.~Novikov \emph{et~al.}, ``Benchmarks for deep off-policy evaluation,'' in \emph{9th {{International Conference}} on {{Learning Representations}}}.\hskip 1em plus 0.5em minus 0.4em\relax {Virtual Event, Austria}: {OpenReview.net}, 2021.

\bibitem{duan2016}
\BIBentryALTinterwordspacing
Y.~Duan, J.~Schulman, X.~Chen, P.~L. Bartlett, I.~Sutskever, and P.~Abbeel, ``{{RL}}$^2$: Fast reinforcement learning via slow reinforcement learning,'' \emph{arXiv:1611.02779 [cs.AI]}, 2016. [Online]. Available: \url{http://arxiv.org/abs/1611.02779}
\BIBentrySTDinterwordspacing

\bibitem{rothfuss2019}
J.~Rothfuss, D.~Lee, I.~Clavera, T.~Asfour, and P.~Abbeel, ``{{ProMP}}: Proximal meta-policy search,'' in \emph{7th {{International Conference}} on {{Learning Representations}}}.\hskip 1em plus 0.5em minus 0.4em\relax New Orleans, LA, USA: OpenReview.net, 2019.

\bibitem{rakelly2019data}
\BIBentryALTinterwordspacing
K.~Rakelly and A.~Zhou, ``{PEARL}: Efficient off-policy meta-learning via probabilistic context variables,'' 2019. [Online]. Available: \url{https://github.com/katerakelly/oyster}
\BIBentrySTDinterwordspacing

\bibitem{Kalashnikov2021}
\BIBentryALTinterwordspacing
D.~Kalashnikov, J.~Varley, Y.~Chebotar, B.~Swanson, R.~Jonschkowski, C.~Finn \emph{et~al.}, ``{{MT-Opt}}: Continuous multi-task robotic reinforcement learning at scale,'' \emph{arxiv:2104.08212 [cs.RO]}, 2021. [Online]. Available: \url{https://arxiv.org/abs/2104.08212}
\BIBentrySTDinterwordspacing

\bibitem{vuong2023}
Q.~Vuong, S.~Levine, H.~R. Walke, K.~Pertsch, A.~Singh, R.~Doshi \emph{et~al.}, ``{Open X-Embodiment}: Robotic learning datasets and {RT-X} models,'' in \emph{Towards Generalist Robots: Learning Paradigms for Scalable Skill Acquisition @ CoRL2023}, Atlanta, Georgia USA, 2023.

\bibitem{fujimoto2018}
S.~Fujimoto, H.~{van Hoof}, and D.~Meger, ``Addressing function approximation error in actor-critic methods,'' in \emph{Proceedings of the 35th {{International Conference}} on {{Machine Learning}}}, J.~Dy and A.~Krause, Eds., vol.~80.\hskip 1em plus 0.5em minus 0.4em\relax {Stockholm, Sweden}: {PMLR}, 2018, pp. 1587--1596.

\bibitem{mnih2015human}
V.~Mnih, K.~Kavukcuoglu, D.~Silver, A.~A. Rusu, J.~Veness, M.~G. Bellemare \emph{et~al.}, ``Human-level control through deep reinforcement learning,'' \emph{Nature}, vol. 518, no. 7540, pp. 529--533, 2015.

\bibitem{lillicrap2016}
T.~P. Lillicrap, J.~J. Hunt, A.~Pritzel, N.~Heess, T.~Erez, Y.~Tassa \emph{et~al.}, ``Continuous control with deep reinforcement learning,'' in \emph{4th {{International Conference}} on {{Learning Representations}}}, Y.~Bengio and Y.~LeCun, Eds., {San Juan, Puerto Rico}, 2016.

\bibitem{haarnoja2018}
\BIBentryALTinterwordspacing
T.~Haarnoja, A.~Zhou, K.~Hartikainen, G.~Tucker, S.~Ha, J.~Tan \emph{et~al.}, ``Soft actor-critic algorithms and applications,'' \emph{arXiv:1812.05905 [cs.LG]}, 2018. [Online]. Available: \url{http://arxiv.org/abs/1812.05905}
\BIBentrySTDinterwordspacing

\end{thebibliography}

\begin{IEEEbiography}[{\includegraphics[width=1in,height=1.25in,clip,keepaspectratio]{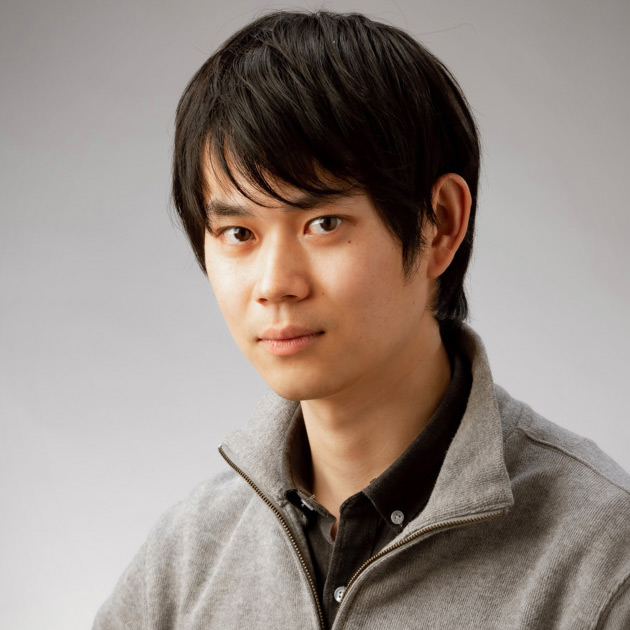}}]{Satoshi Yamamori}
received the B.E. and M.I. degrees from Kyoto University, in 2017 and 2019, respectively.
He was a Research Fellow (DC1) at Japan Society for the Promotion of Science (JSPS) from 2019 to 2022.
He works with the Advanced Telecommunications Research Institute International (ATR).
He is also affiliated with the Graduate School of Informatics, Kyoto University, Kyoto.
His research interests include reinforcement learning, robot control, and learning machine systems.
\end{IEEEbiography}
\begin{IEEEbiography}[{\includegraphics[width=1in,height=1.25in,clip,keepaspectratio]{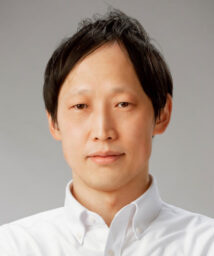}}]{Jun Morimoto}
received the Ph.D.
degree in information science from the Nara Institute
of Science and Technology, Nara, Japan, in 2001.
From 2001 to 2002, he was a Post-Doctoral Fellow with The Robotics Institute, Carnegie Mellon
University, Pittsburgh, PA, USA. He joined the
Advanced Telecommunications Research Institute
International (ATR), Kyoto, Japan, in 2002. He is currently a Professor
with the Graduate School of Informatics, Kyoto University, Kyoto. He is
also the Head of the Department of Brain Robot Interface (BRI), ATR
Computational Neuroscience Laboratories, and a Senior Visiting Scientist
of the Man-Machine Collaboration Research Team, Guardian Robot Project,
RIKEN, Kyoto
\end{IEEEbiography}

\end{document}